\documentclass[]{fairmeta}

\usepackage{wrapfig}
\usepackage{tabularx}
\usepackage{textcomp}
\usepackage{stfloats}
\usepackage{url}
\usepackage{verbatim}
\usepackage{titlesec}
\usepackage{tocloft}
\usepackage{adjustbox}
\usepackage{tikz}
\usepackage{comment}
\usepackage{amsmath,amssymb}
\usepackage{colortbl}
\usepackage{color}
\usepackage{hyperref}
\usepackage{graphicx}
\usepackage{subcaption} 
\usepackage{multirow}
\usepackage{booktabs}
\usepackage{xcolor}
\usepackage{pifont}
\usepackage{bbding}
\usepackage{fontawesome}
\usepackage{xspace}
\usepackage{float}
\usepackage{algorithm}
\usepackage{listings}
\usepackage{algcompatible}
\usepackage{algpseudocode}
\usepackage{soul}
\usepackage{array}
\usepackage{tcolorbox}
\usepackage{diagbox}
\usepackage{makecell}
\usepackage{rotating}
\usepackage{enumitem}
\usepackage{natbib}

\usepackage{amsmath,amsfonts,bm}

\def\eqref#1{equation~\ref{#1}}

\def\1{\bm{1}}

\DeclareMathAlphabet{\mathsfit}{\encodingdefault}{\sfdefault}{m}{sl}
\SetMathAlphabet{\mathsfit}{bold}{\encodingdefault}{\sfdefault}{bx}{n}

\RequirePackage{xspace}
\makeatletter
\DeclareRobustCommand\onedot{\futurelet\@let@token\@onedot}
\def\@onedot{\ifx\@let@token.\else.\null\fi\xspace}

\makeatother

\definecolor{adptorange}{RGB}{248, 205, 172}
\definecolor{cmpblue}{RGB}{189, 215, 238}
\definecolor{cmpblue}{RGB}{189, 215, 238}

\definecolor{our_red}{RGB}{232,157,160}
\definecolor{our_blue}{RGB}{136,206,230}
\definecolor{our_orange}{RGB}{246,200,168}
\definecolor{our_green}{RGB}{178,211,164}

\definecolor{attn_code0}{RGB}{247,215,200}
\definecolor{attn_code1}{RGB}{238,169,139}
\definecolor{mlp_code0}{RGB}{204,201,221}
\definecolor{mlp_code1}{RGB}{102,95,153}

\definecolor{token_blue}{RGB}{84, 120, 140}

\newlength\savewidth

\newcolumntype{x}[1]{>{\centering\arraybackslash}p{#1pt}}
\newcolumntype{y}[1]{>{\raggedright\arraybackslash}p{#1pt}}
\newcolumntype{z}[1]{>{\raggedleft\arraybackslash}p{#1pt}}

\renewcommand{\paragraph}[1]{\vspace{1mm}\noindent\textbf{#1}}

\renewcommand{\paragraph}[1]{\vspace{1.25mm}\noindent\textbf{#1}}

\definecolor{codeblue}{rgb}{0.25, 0.5, 0.5}
\definecolor{codekw}{rgb}{0.35, 0.35, 0.75}
\lstdefinestyle{Pytorch}{
    language = Python,
    backgroundcolor = \color{white},
    basicstyle = \fontsize{9pt}{8pt}\selectfont\ttfamily\bfseries,
    columns = fullflexible,
    aboveskip=1pt,
    belowskip=1pt,
    breaklines = true,
    captionpos = b,
    commentstyle = \color{codeblue},
    keywordstyle = \color{codekw},
}

\definecolor{green}{HTML}{009000}
\definecolor{red}{HTML}{ea4335}

\newcommand{\correspond}{\spadesuit}

\newcommand{\ours}{GP3 \xspace}

\newtcolorbox{promptblock}{
    colback=gray!5,
    colframe=gray!15,
    boxrule=0.5pt,
    arc=3pt,
    left=12pt,
    right=12pt,
    top=8pt,
    bottom=8pt,
    boxsep=8pt,
    breakable
}

\title{\ours: A 3D Geometry-Aware Policy with Multi-View Images for Robotic Manipulation}

\author{
Quanhao Qian$^{1,2}$, 
Guoyang Zhao$^{1,3}$, 
Gongjie Zhang$^{1,2}$, 
Jiuniu Wang$^{1,2}$, 
Ran Xu$^{1,2\correspond}$,
Junlong Gao$^{1,2}$, 
Deli Zhao$^{1,2}$
}

\affiliation{$^{1}$DAMO Academy, Alibaba Group 

$^{2}$HuPan Lab
$^{3}$Tongji University
}

\contribution[\correspond]{Corresponding author}

\abstract{
Effective robotic manipulation relies on a precise understanding of 3D scene geometry, and one of the most straightforward ways to acquire such geometry is through multi-view observations. Motivated by this, we present GP3---a 3D geometry-aware robotic manipulation policy that leverages multi-view input. GP3 employs a spatial encoder to infer dense spatial features from RGB observations, which enable the estimation of depth and camera parameters, leading to a compact yet expressive 3D scene representation tailored for manipulation. This representation is fused with language instructions and translated into continuous actions via a lightweight policy head. Comprehensive experiments demonstrate that GP3 consistently outperforms state-of-the-art methods on simulated benchmarks. Furthermore, GP3 transfers effectively to real-world robots without depth sensors or pre-mapped environments, requiring only minimal fine-tuning. These results highlight GP3 as a practical, sensor-agnostic solution for geometry-aware robotic manipulation.

}

\date{\today}
\correspondence{Ran Xu, \texttt{xr118482@alibaba-inc.com}}

\begin{document}
\thispagestyle{firstheader}
\maketitle
\pagestyle{fancy}
\fancyhf{}
\fancyfoot[C]{\thepage}

\section{Introduction}
\label{sec:intro}

Spatial perception is a core capability for general-purpose robotic motion, particularly in enabling robots to fully perceive and utilize 3D geometric information in the scene. A dominant line of research integrates 3D information into visuomotor policies by directly encoding point cloud data~\citep{chen2024sugarpretraining3dvisual, wilcox2025adapt3radaptive3dscene, ke2023d_3da, ze20243d_3dp, Yang2025FP3A3, Zhu2024PointCM, jia2024lift3d, huang2024imagination, haldar2025point, eisner2022flowbot3d, liu2024voxactbvoxelbasedactingstabilizing, grotz2024peract2,miao2025towards, ravan2024combiningplanningdiffusionmobility}. However, these methods depend on specialized hardware like depth sensors, which can be unreliable or unavailable in many real-world scenarios. Conversely, approaches that generate implicit 3D representations from standard RGB images~\citep{gervet2023act3d, xian2023chaineddiffuser, zhu2025spa, chen2025robohorizonllmassistedmultiviewworld, seo2023multi, qian20243d, zhang2023online} often struggle with generalization, producing spatial representations that are not robust enough for diverse, unseen environments.
\begin{figure*}[!htp]
\centering
\includegraphics[width=0.9\textwidth]{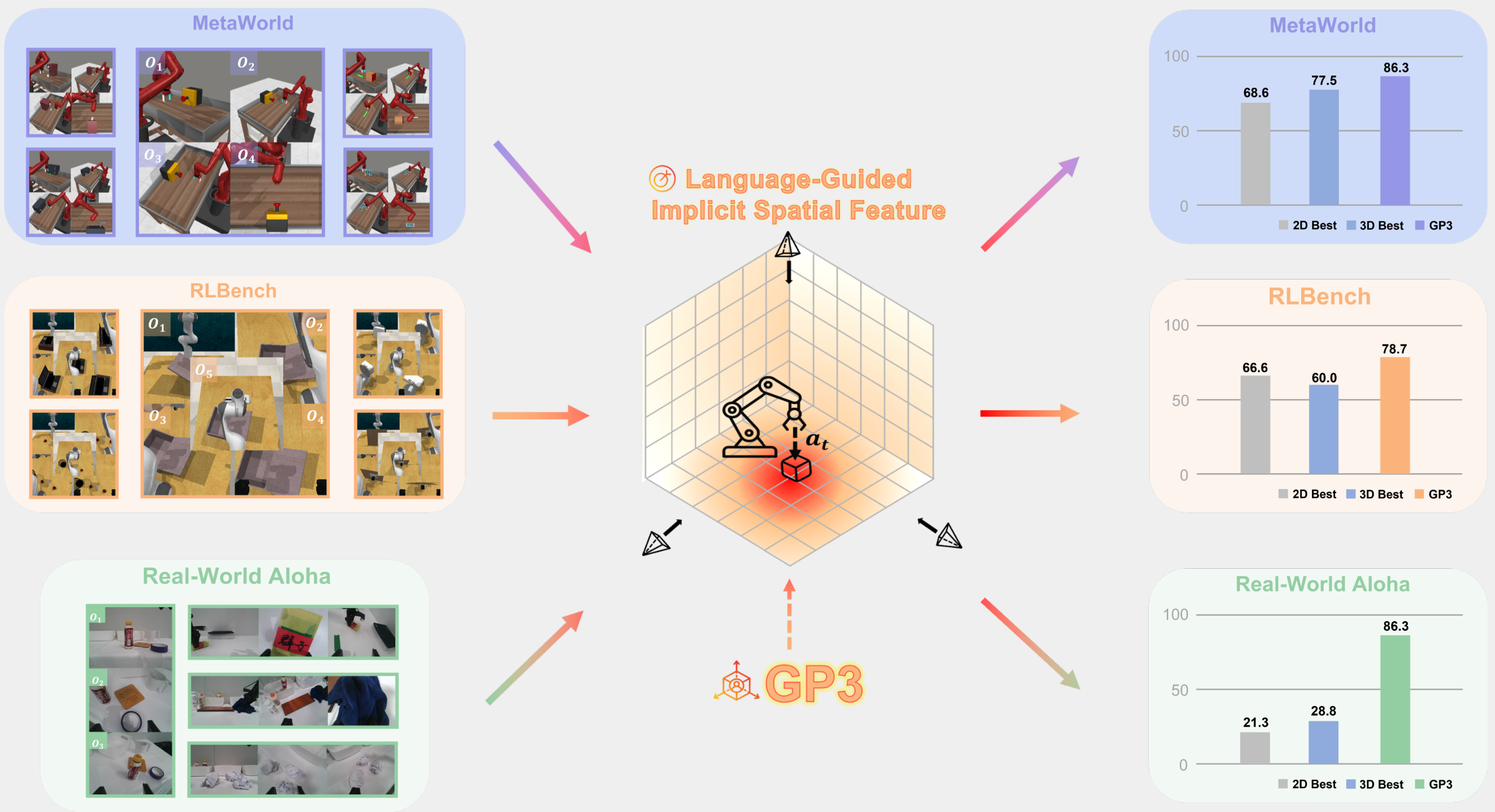}
\caption{
We introduce GP3, a 3D geometry‑aware policy framework for robotic manipulation that extracts implicit spatial features from multi‑view images with language guidance. Compared to both 2D‑based and 3D‑based baseline methods, GP3 consistently outperforms them on simulated and real‑world manipulation tasks under same settings.}
\label{fig:first_pic}
\end{figure*}

To bridge this gap, we introduce GP3, a geometry-aware policy for robotic manipulation that achieves robust multi-view spatial reasoning without requiring depth data. GP3 is built upon two key technical contributions: a robot-adapted spatial understanding module and an efficient attention mechanism for multi-view, language-guided control.

First, the core of GP3 is RoboVGGT, a spatial encoder adapted for robotic tasks. We begin with a large-scale pretrained 3D reconstruction model, VGGT~\citep{wang2025vggt}, selected for its exceptional generalization capabilities across diverse scenes. We then finetune it on a newly curated, multi-domain robotics dataset, comprising simulated data from RLBench~\citep{james2020rlbench}, MetaWorld~\citep{yu2020meta}, and RoboTwin~\citep{mu2025robotwin}, alongside real-world task data. This targeted finetuning strengthens RoboVGGT with generalizable spatial reasoning abilities grounded in multi-view geometry, enabling accurate 3D understanding across a wide spectrum of robotic scenes and tasks.

Second, as evidenced in Section~\ref{section:E_3}, we observed that simply increasing the number of input views can paradoxically degrade performance by introducing distracting information and diluting the model's focus. To counteract this, we introduce G-FiLM (Global attention-based Feature-wise Linear Modulation). Inspired by FiLM~\citep{perez2018film}, G-FiLM integrates language instructions to dynamically guide the policy's attention. This allows the model to focus on task-relevant spatial features from the multi-view input while actively suppressing irrelevant noise, significantly improving task success.

Finally, our experimental results demonstrate that the proposed approach eliminates the need for explicit depth sensors while retaining robust 3D reconstruction capabilities for previously unseen robotic tasks. The method efficiently adapts to new environments using only minimal amounts of data, and consistently outperforms prior methods reliant on explicit 3D information~\citep{jia2024lift3d, ze20243d_3dp} as well as other approaches with RBG inputs~\citep{zhu2025spa, radford2021learning_clip, majumdar2023we-vc1, nair2022rm, chi2023diffusion}. Under the same implementation settings, GP3 improves over the best baseline by 11.2\% on MetaWorld, by 22.7\% on RLBench, and by 57.5\% in real‑world experiments.

In summary, our contributions are threefold:
\begin{itemize}
\item We introduce \textbf{RoboVGGT}, a geometry-aware 3D reconstruction model fine-tuned on a curated dataset combining simulated and real-world data, achieving robust and generalizable 3D reconstruction across diverse robotic scenarios.
\item We present \textbf{G-FiLM}, a global attention-based modulation mechanism that facilitates task-relevant attention while suppressing redundant information in multi-view perception.
\item With the above two key technical contributions, we propose \textbf{GP3}, a 3D geometry-aware policy for robotic manipulation that enables robust and efficient multi-view spatial reasoning. GP3 achieves state-of-the-art performance across multiple simulated and real-world benchmarks, setting a new standard for robust visuomotor control without relying on specific sensors.
\end{itemize}

\begin{figure*}[t]
\centering
\includegraphics[width=0.9\textwidth]{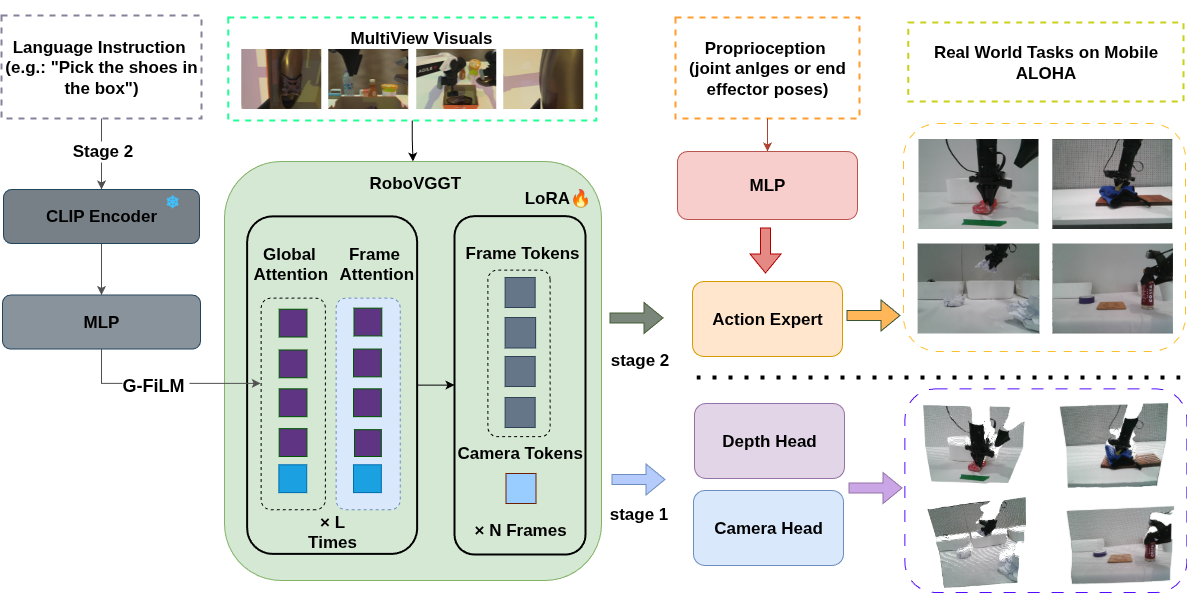}
\caption{
Architecture overview. Our framework adopts a two-stage training pipeline. 
Stage~1 (\textit{geometry model fine-tuning}): RoboVGGT is fine-tuned by equipping it with a multi-view camera parameter head and a depth estimation head. 
Stage~2 (\textit{action prediction training}): We incorporate global attention-based feature-wise linear modulation (G-FiLM), enabling the encoder to focus on task-relevant regions of interest (ROIs) and suppress task-irrelevant noise through adaptive feature modulation. The extracted spatial features are then combined with proprioceptive information to predict the action.
}
\label{fig:brief}
\end{figure*}
\section{Related Work}
\label{sec:related}
{\bfseries Robotic Manipulation. }Traditional robotic manipulation has largely relied on state-based reinforcement learning~\citep{lillicrap2015continuous, li2017deep, schulman2017proximal}, assuming full observability of low-dimensional state representations (e.g., joint angles, end-effector poses). Although effective in simulation and controlled environments, these methods are challenging to deploy in real-world scenarios due to the difficulty of accurate state estimation and the high cost of specialized sensing hardware~\citep{gu2017deep}.

Recent advances~\citep{jang2022bc, brohan2022rt, shang2024theiadistillingdiversevision, mazzia2022action, chi2023diffusion} have shifted towards learning control policies directly from visual observations, often leveraging large-scale datasets and modern architectures such as Vision Transformers (ViTs)~\citep{shridhar2021cliportpathwaysroboticmanipulation}. For example, R3M~\citep{nair2022rm} employs contrastive learning to obtain universal embodied representations from diverse human video data, while VC-1~\citep{majumdar2023we-vc1} evaluates the effectiveness of masked auto-encoding (MAE)~\citep{He2021MaskedAA} strategies for robotic pretraining. Vision-Language-Action (VLA) models~\citep{kimopenvla, brohan2023rt2visionlanguageactionmodelstransfer, liu2024robomambaefficientvisionlanguageactionmodel} further unify perception, language grounding, and action prediction within a single framework, improving task generalization but still facing challenges in the fine-grained spatial reasoning required for precision tasks such as assembly or insertion~\citep{fang2023anygrasp}.   
This limitation has motivated growing interest in methods that explicitly incorporate spatial and geometric cues into robotic perception and control.

{\bfseries Spatial Geometry-Based Methods. }
To address the limitations of 2D based method in capturing precise spatial relationships, recent research has explored integrating 3D geometric information into robotic perception and control. Such approaches leverage depth, point clouds, or multi-view geometry to enhance spatial reasoning and improve manipulation accuracy, particularly in tasks requiring fine-grained positioning and contact reasoning. 3D-aware robotic perception research can be broadly categorized into two complementary paradigms--explicit and implicit geometry representations.

\textbf{Explicit Geometry Based Approaches.} Recent advancements in robotic perception have predominantly relied on explicit spatial geometry representations, including point clouds, voxels, and Gaussian splats for 3D spatial reasoning. Representative frameworks such as 3DP~\citep{ze20243d_3dp}, 3DA~\citep{ke2023d_3da}, LIFT3D~\citep{jia2024lift3d}, and RVT~\citep{goyal2023rvt, goyal2024rvt} demonstrate the effectiveness of point cloud-based approaches, while voxel-based methods like PerACT~\citep{grotz2024peract2, shridhar2022perceiveractormultitasktransformerrobotic}, and VoxAct~\citep{liu2024voxactbvoxelbasedactingstabilizing} further expand this paradigm. However, these methods often require expensive depth sensors and substantial computation, and their performance is further hindered by occlusions, sensor noise, and the lack of real-world evaluation beyond curated benchmarks. To address these drawbacks, recent works, including our proposed framework, explore RGB-only alternatives that retain geometric reasoning capabilities without relying on explicit depth.

\textbf{Implicit Geometry-Based Approaches.} Inspired by multi-modal alignment approaches~\citep{zhai2023sigmoidlosslanguageimage, radford2021learning_clip}, recent works such as~\citep{pang2024depth, wang2024visualroboticmanipulationdepthaware} attempt to address depth dependency through depth-augmented image features. While these models can predict depth from visual input, single-frame depth estimation remains inherently ambiguous due to missing scale information and absolute depth resolution without additional constraints. To overcome this limitation, approaches like RoboHorizon~\citep{chen2025robohorizonllmassistedmultiviewworld} and 3D-MVP~\citep{qian20243d} employ multi-view MAE to learn implicit 3D information of a robot workspace. The most relevant work to ours is SPA~\citep{zhu2025spa}, which enhances ViT with 3D awareness through differentiable neural rendering on multi-view images. However, the volumetric features produced by SPA predominantly support coarse-grained spatial reasoning, which is insufficient for precision-critical robotic tasks. In addition, these previous methods lack generalization capability and fail to understand spatial relationships in unseen manipulation tasks or scenarios. In this paper, we propose a pixel-wise spatial reasoning framework that enables fine-grained perception crucial for complex robotic applications by transferring the pretrained geometry transformer to robotic tasks.  

{\bfseries Transformer-Based 3D Vision Methods. }Classical scene reconstruction methods~\citep{BAreconstruction, reconstruction2, colmap_Schonberger_2016_CVPR, pan2024globalstructurefrommotionrevisited} leverage multi-view geometry to recover 3D environments. A newer trend centers on transformer-based foundation models~\citep{Wang20243DRW, wang2024dust3r, leroy2024grounding_mast3r, wang2025continuous, yang2025fast3r, Smart2024Splatt3RZG, wang2025vggt}, which reconstruct dense scenes from raw RGB, even when camera parameters are unknown. DUSt3R~\citep{wang2024dust3r} and MASt3R~\citep{leroy2024grounding_mast3r} pioneered joint geometry and camera estimation, while CUT3R~\citep{wang2025continuous} and Fast3R~\citep{yang2025fast3r} improved reconstruction speed. VGGT~\citep{wang2025vggt} currently achieves state-of-the-art accuracy and stability in local 3D mapping.

Our model, RoboVGGT, builds upon VGGT’s pretrained capacity for geometry extraction and adapts it to robotic manipulation, where performance in object-centric, small-workspace environments is critical. Since the generalization ability of transformer-based 3D models is highly sensitive to dataset diversity, we collect targeted robotic datasets and fine-tune RoboVGGT with both geometry-specific and task-specific objectives. Furthermore, we integrate language conditioning into the pipeline, enabling the extraction of high-quality, 3D-aware features specialized for robotic manipulation.

\section{Methods}
\label{sec:training}

We aim to model 3D visual geometry for robotic manipulation tasks without relying on explicit dense 3D inputs. As illustrated in Figure~\ref{fig:brief}, we introduce a spatial encoder that extracts rich visual–geometric features from multiple RGB views and passes them to the policy head through a two-stage training.  

Section~\ref{section:M_A} defines the problem, Section~\ref{section:M_B} describes the training of the spatial encoder, Section~\ref{section:M_C} introduces the fusion of language and spatial features by G-FiLM, and Section~\ref{section:M_D} provides a detailed explanation of the action training process.

\subsection{Problem definition and notation}~\label{section:M_A}
The inputs consist of a sequence of RGB images $I_i$ captured from either onboard or third-person cameras, the robot's proprioceptive state $P$, and a language instruction $L$. The objective is to predict the corresponding sequence of robot actions.

\subsection{Geometry-Aware Spatial Encoder}
\label{section:M_B}

To capture 3D geometric information in multi-view inputs, we employ a geometry-aware spatial encoder, thus avoiding the dependency on point clouds as in~\citep{jia2024lift3d}. We build the RoboVGGT based on VGGT~\citep{wang2025vggt} for its state-of-the-art 3D performance; unlike methods that require explicit 3D priors, VGGT inherently learns inter-frame correspondences and reconstructs scenes from multi-view RGB images.  

\begin{figure}[!htb]
    \centering
    \subfloat[VGGT on Sim]{
        \includegraphics[width=0.22\columnwidth]{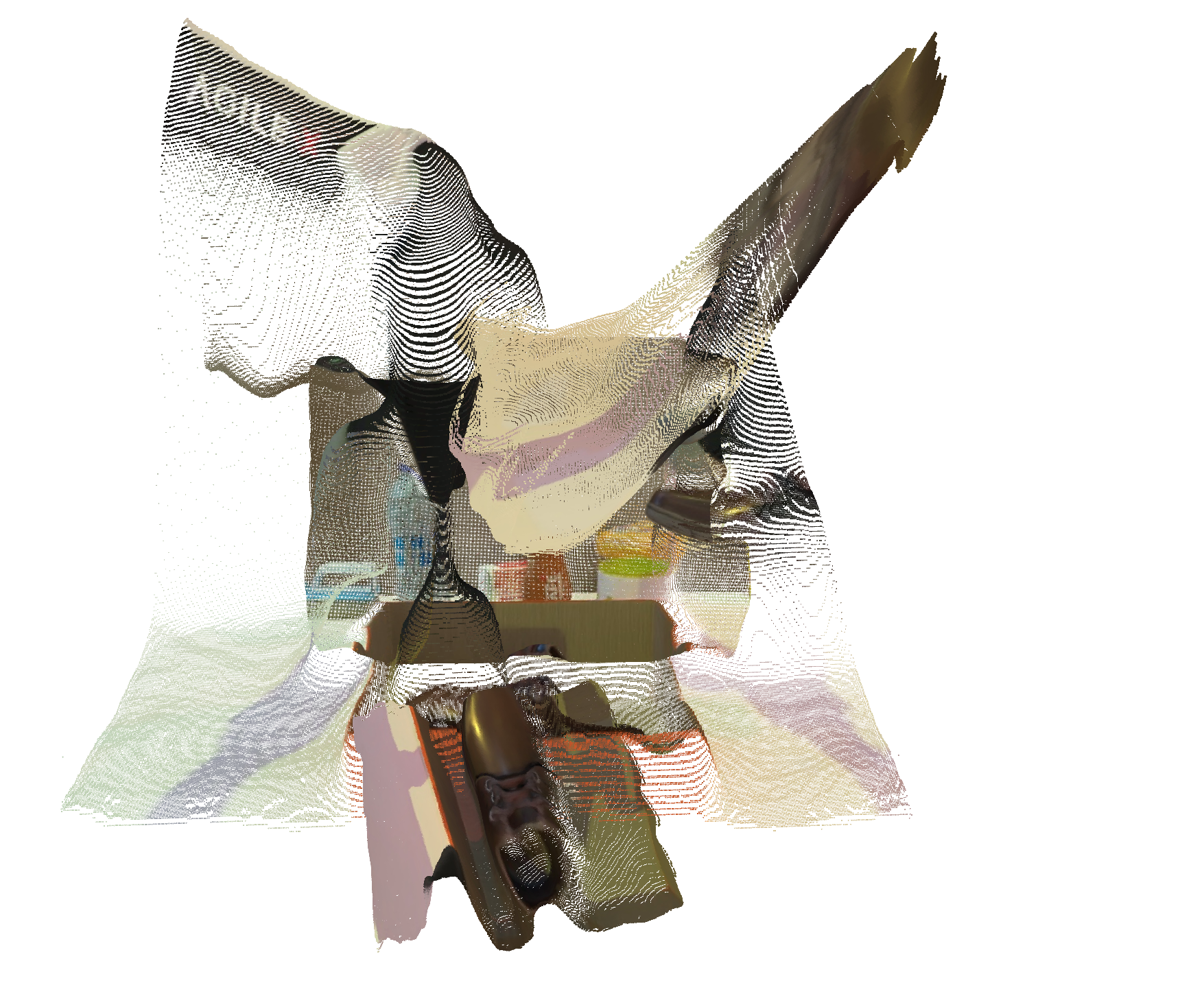}
        \label{fig:sub1}
    }
    \hfill
    \subfloat[RoboVGGT on Sim]{
        \includegraphics[width=0.22\columnwidth]{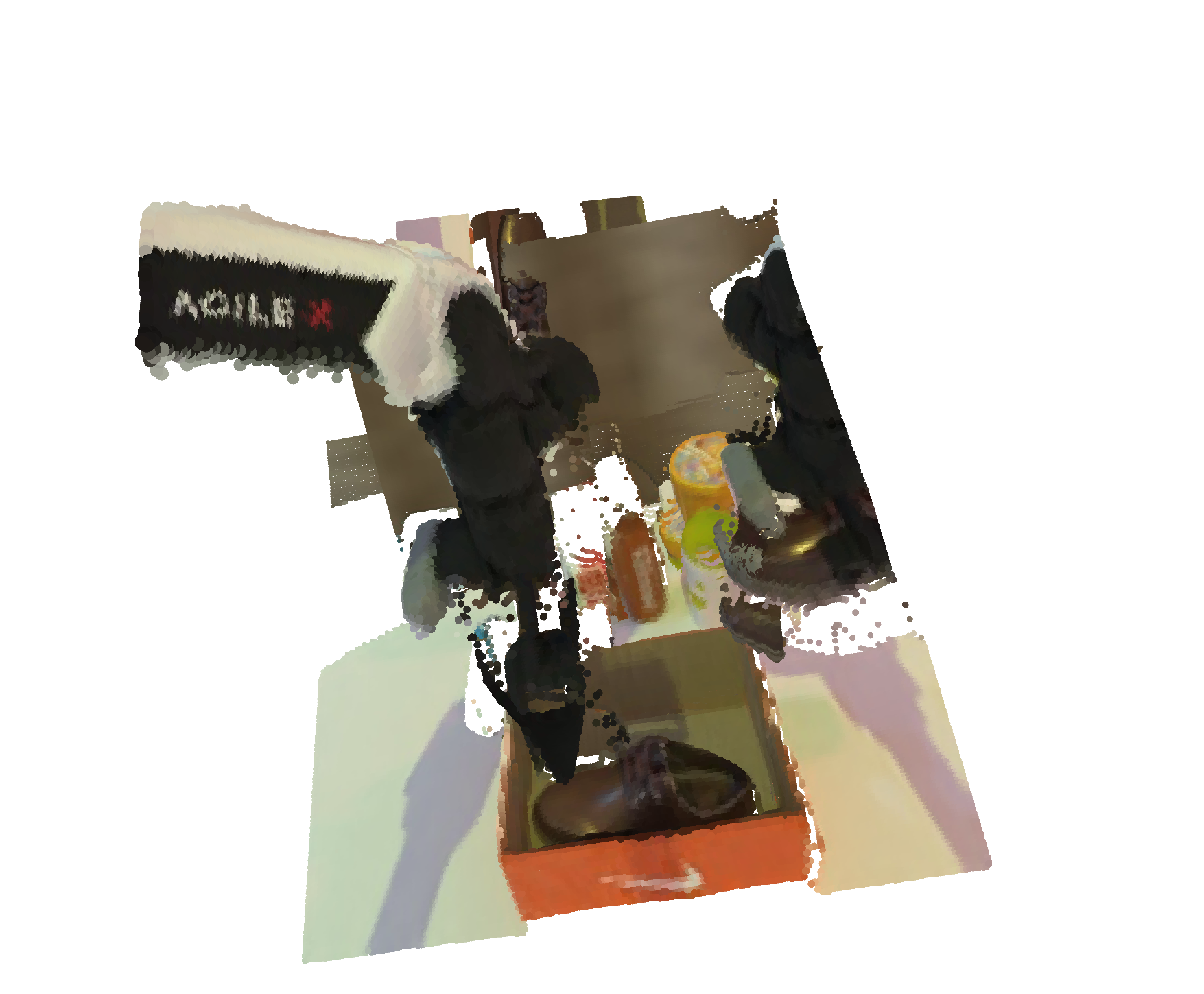}
        \label{fig:sub2}
    }
    \hfill
    \subfloat[VGGT on Real]{
        \includegraphics[width=0.22\columnwidth]{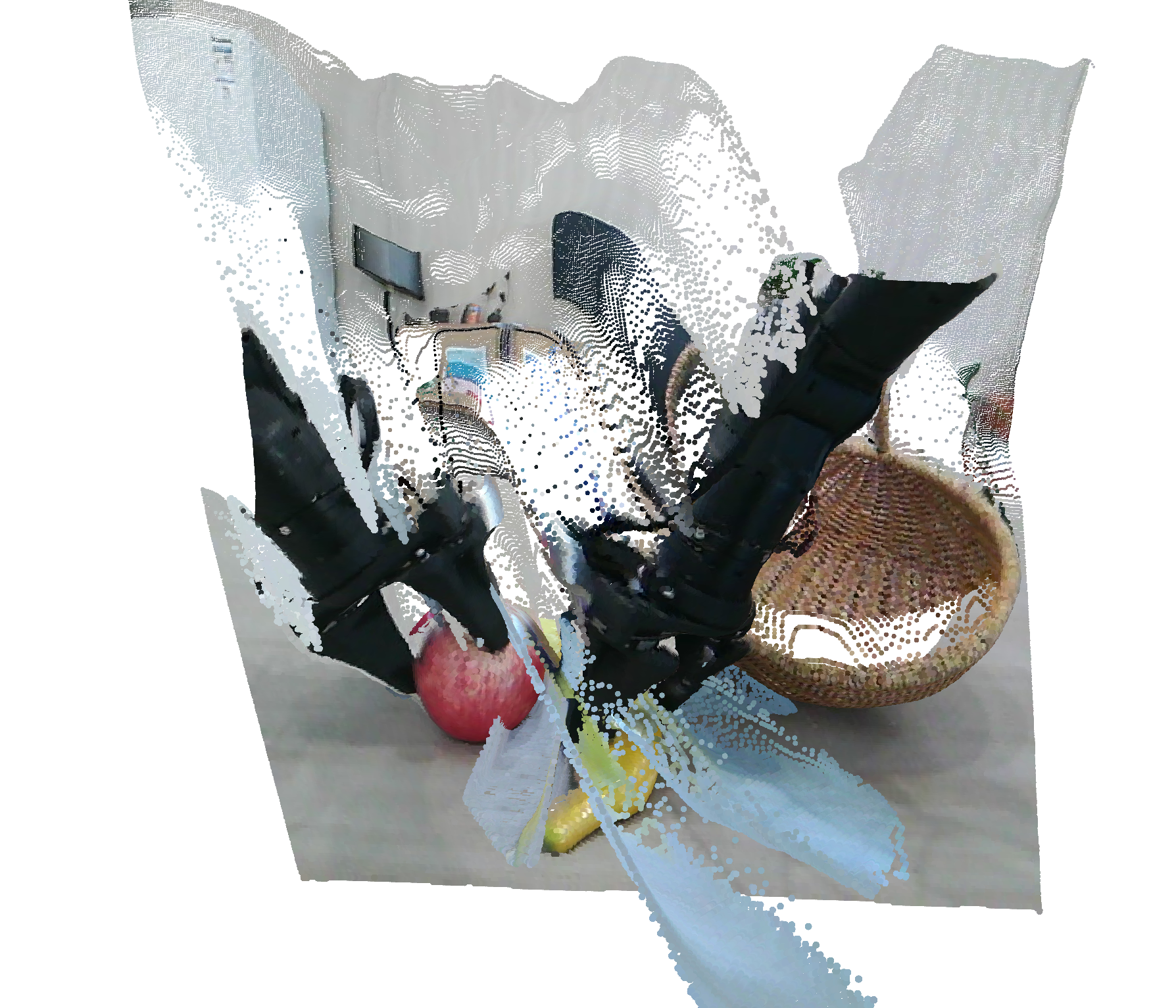}
        \label{fig:sub3}
    }
    \hfill
    \subfloat[RoboVGGT on Real]{
        \includegraphics[width=0.22\columnwidth]{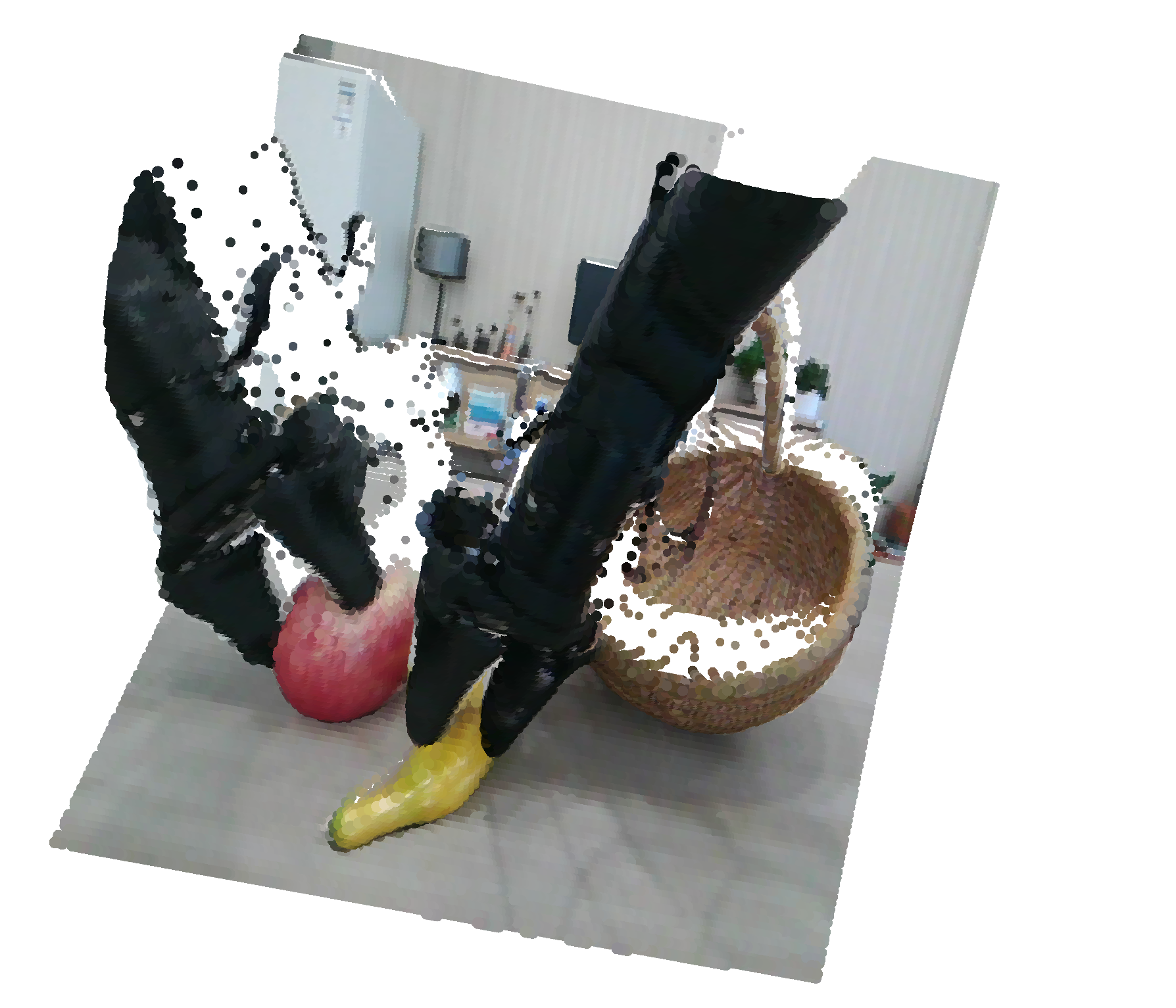}
        \label{fig:sub4}
    }
    \caption{Reconstruction results comparing the original VGGT model against our fine-tuned RoboVGGT model, using both simulated and real-world inputs. Our targeted fine-tuned spatial encoder shows clear improvement on robot manipulation scenes.}
    \label{fig:robotwins_pretrain}
\end{figure}
VGGT's original design for high-resolution inputs leads to slow training/inference and high memory costs. To mitigate this computational bottleneck and bridge the domain gap between its pretraining data and robotic manipulation, we fine-tune the model at a practical $224 \times 224$ resolution. Our training leverages 150 K simulated frames from RLBench~\citep{james2020rlbench}, MetaWorld~\citep{yu2020meta} and RoboTwins~\citep{mu2025robotwin}, along with 20 K real-world frames, and incorporates explicit geometric supervision to improve adaptation.

We follow the VGGT framework to train the spatial encoder using a multi-task camera and depth loss:
\begin{equation}
\mathcal{L} = \mathcal{L}_{\text{camera}} + \mathcal{L}_{\text{depth}} .
\end{equation}

The camera loss $\mathcal{L}_{\text{camera}}$ supervises the predicted camera parameters $g_i$ against the ground-truth parameters $g^{gt}_i$ with Huber loss $\| \cdot \|_{\epsilon}$:
\begin{equation}
\mathcal{L}_{\text{camera}} = \sum_{i=1}^N \| g^{gt}_i - g_i \|_{\epsilon}.
\end{equation}
The depth loss $\mathcal{L}_{\text{depth}}$ integrates aleatoric-uncertainty weighting with gradient-based depth supervision:
\begin{equation}
\mathcal{L}_{\text{depth}}
= \sum_{i=1}^N
(\| \Sigma_D^i \odot (D^{gt}_i - D_i) \|+ \| \Sigma_D^i \odot (\nabla D^{gt}_i - \nabla D_i) \| - \alpha \log \Sigma_D^i ),
\end{equation}
where $\Sigma_D^i$ is the predicted per-pixel uncertainty map~\citep{kendall2017uncertainties}, $\odot$ denotes element-wise multiplication, $D^{gt}_i$ is the ground-truth depth map, $D_i$ is the predicted depth map, $\nabla$ represents the spatial gradient operator, and $\alpha$ is a weighting coefficient controlling the uncertainty regularization.

As shown in Figure~\ref{fig:robotwins_pretrain}, we input the multi-view images captured by front camera and the wrist cameras to spatial encoder, original model lacks the ability to generalize to robotic scenes, exhibiting particularly poor reconstruction quality for the robotic arm. After our fine-tuning process, the model exhibits robust generalization to reconstruct robotic manipulation scenes. Notably, the robotic task in the depicted real-world scene was unseen by the model during the training process, further demonstrating the generalization ability of our fine-tuned model to robotic scenes.

\begin{figure}[!htb]
    \centering
    \subfloat[VGGT with 224$\times$224 input] {
    \includegraphics[width=0.3\columnwidth]{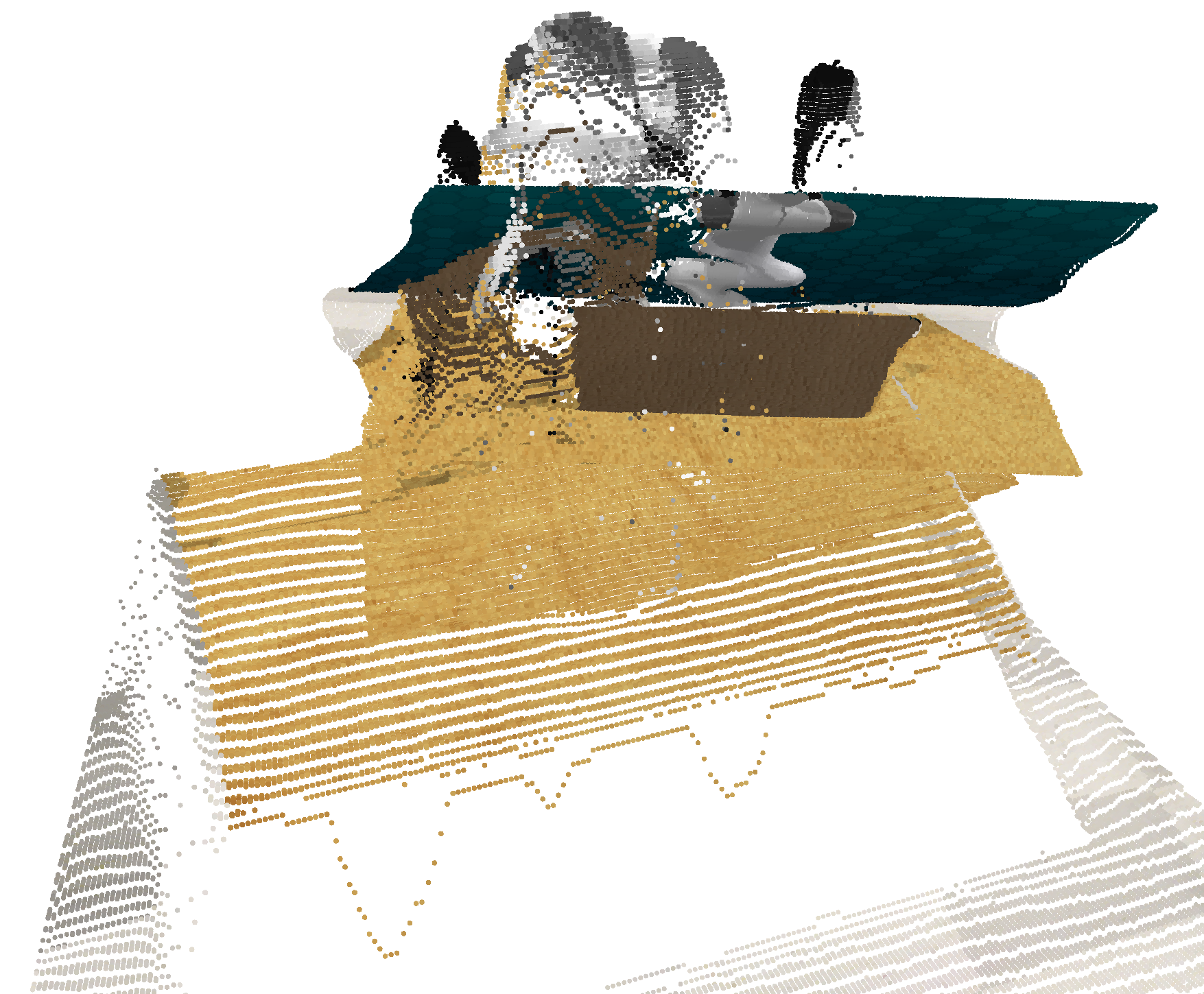}
    }
    \hspace{0.05\columnwidth} 
    \subfloat[RoboVGGT with 224$\times$224 input] {
    \includegraphics[width=0.3\columnwidth]{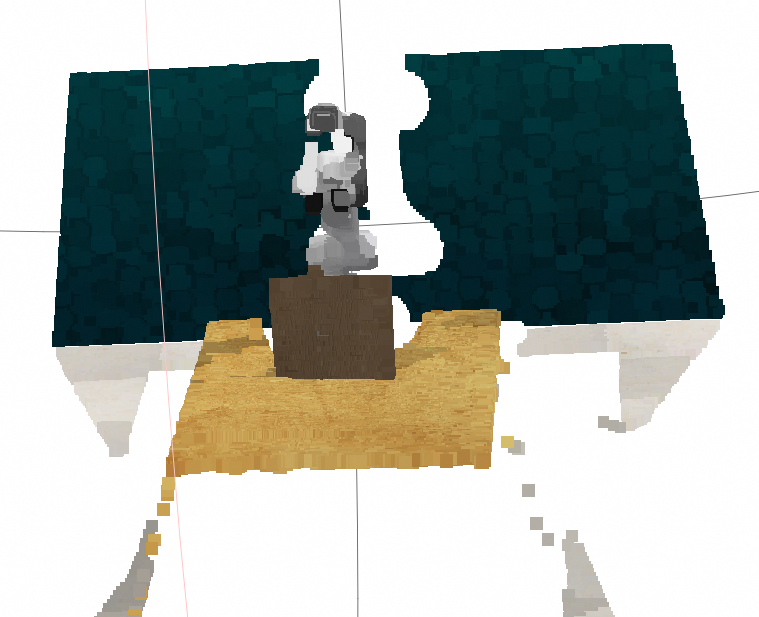}
    }
\caption{The outcome of reconstructing VGGT using a resolution of 224$\times$224, despite the fact that VGGT without fine-tuning is unable to process such a resolution.} 
    \label{fig:vggt_pretrain}
\end{figure}

As demonstrated in Figure~\ref{fig:vggt_pretrain}, the off-the-shelf VGGT model struggles to reconstruct multi-view scenes at a resolution of 224$\times$224 pixels. However, after minimal fine-tuning, RoboVGGT's performance at this lower resolution becomes comparable to its performance at a resolution 518$\times$518 pixels.

\begin{figure}[!htb]
    \centering
    \subfloat[Observations] {
    \includegraphics[width=0.9\columnwidth]{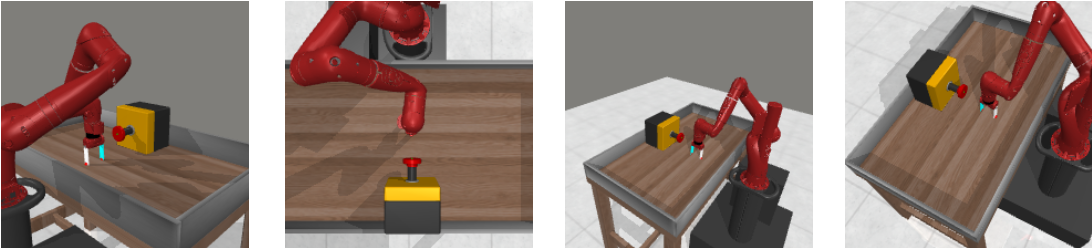}
    }
    \hspace{0.05\columnwidth} 
    \subfloat[G-FiLM's Attention Responses] {
    \includegraphics[width=0.9\columnwidth]{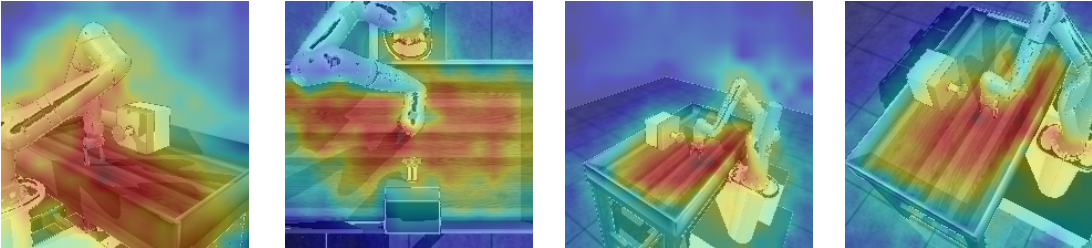}
    }
\caption{Our proposed G-FiLM effectively guides attention to the visual contents pertinent to the button-press task.} 
    \label{fig:intro}
\end{figure}

\subsection{Global Attention-based Feature-wise Linear Modulation}~\label{section:M_C}
While the spatial encoder can extract implicit spatial information from multi-frame inputs, we observe that increasing the number of views does not always improve task success rates and may even degrade performance. This is primarily due to the redundant and task-irrelevant information introduced in multi-view settings.  
The OpenVLA-oft~\citep{kim2025fine_openvla-oft} framework addresses this by applying FiLM~\citep{perez2018film} modules to the self-attention components of each vision transformer block.

In our work, the transformer architecture integrates both global and frame-wise attention layers, with global attention designed to capture cross-perspective associations, while frame-wise attention focuses primarily on local image feature extraction. To address this, we propose G-FiLM, a novel adaptation that applies FiLM modulation exclusively to global attention patterns. Our formulation is defined as:
\begin{equation}
\text{G-FiLM}(F) =  \gamma \odot (M * F) + \beta,
\end{equation}

where $\gamma$ and $\beta$ are scaling and shifting vectors projected by language embeddings, modulate the visual features $F$ through an affine transformation. The operator $\odot$ denotes element-wise multiplication. 
$M$ is a mask where each element $M_j$ corresponds to an attention layer $j$.
\begin{equation}
M_j = \begin{cases} \text{True} & \text{if } j \text{ is a global attention layer.} \\ \text{False} & \text{otherwise.} \end{cases}
\end{equation}

This formulation enables semantic-aware modulation of attention patterns while maintaining computational efficiency through parameter sharing across feature dimensions. The corresponding experimental results are summarized in ablation Table~\ref{tab:ablation}, showing consistent performance improvements over all  baseline methods.
Figure~\ref{fig:intro} displays the visualization results of task focus, obtained by normalizing the global attention token, proving that G-FiLM aids the model in focusing on the areas pertinent to button-pressing task.

\subsection{Action Training}~\label{section:M_D}
For each robot, the action representation is selected to match its control interface: we adopt the end-effector pose for simulated tasks and the joint-space configuration for the real-world ALOHA system. The action loss is defined as

\begin{equation}
\mathcal{L}_{\text{action}} = \sum{MSE(A^{gt} - A)}.
\end{equation}

Specifically, we leverage LoRA~\citep{hu2022lora} for efficient fine-tuning of the VGGT encoder weights throughout all training stages including action training. 
Let $W_0 \in \mathbb{R}^{d \times k}$ denote a frozen pre-trained weight matrix from the VGGT encoder. 
LoRA introduces a trainable low-rank decomposition to model the weight update as:
\begin{equation}
W' = W_0 + \Delta W, \quad \Delta W = \frac{\alpha}{r} B A ,
\end{equation}
where $A \in \mathbb{R}^{k \times r}$ and $B \in \mathbb{R}^{r \times d}$ are learnable matrices, 
$r \ll \min(d,k)$ is the rank of the adaptation, and $\alpha$ is a scaling factor controlling the update magnitude. 
During fine-tuning, only $A$ and $B$ are updated while $W_0$ remains frozen, 
thus enabling parameter-efficient training.
\section{Experiments}
\label{sec:eval_framework}

\subsection{Simulation Experiment}~\label{section:E_1}
{\bfseries Benchmarks. }We evaluate GP3 on diverse tasks drawn from two widely used robotic manipulation simulation benchmarks: MetaWorld~\citep{yu2020meta}, running in the MuJoCo simulator, and RLBench~\citep{james2020rlbench}, running in the CoppeliaSim simulator.  
In MetaWorld, which features a tabletop environment with a Sawyer robotic arm and a two-finger gripper, we select 50 tasks of varying difficulty levels. In RLBench, which uses a Franka Panda robot equipped with multi-view cameras, we evaluate on the same six tasks as~\citep{jia2024lift3d}.

{\bfseries Data collection. }In MetaWorld, we generate 25 expert demonstrations per task using scripted policies; each trajectory terminates immediately upon task success, unlike~\citep{ze20243d_3dp}, which records a fixed 200-step horizon, we collect images from three corner cameras and one overhead camera.  
For RLBench, demonstration trajectories are generated via predefined waypoints using the Open Motion Planning Library~\citep{open_motion}, resulting in 100 training demonstrations per task. We collect images from front camera, overhead camera and wrist camera.

{\bfseries Data Baselines. }GP3 systematically enhances multi-view representations for robotic manipulation. To assess its contribution, we benchmark it against 7 representative methods, grouped in 3 categories:
\textbf{(1) 2D Robotic Representations Methods}: 2D foundation model CLIP (ViT-B/16)~\citep{radford2021learning_clip} and the 2D pretrained policy backbone R3M~\citep{nair2022rm}, VC-1~\citep{majumdar2023we-vc1}.
\textbf{(2) Explicit 3D Policies}: DP3~\citep{ze20243d_3dp} and pretrained 3D policy Lift3D~\citep{jia2024lift3d}
\textbf{(3) Implicit 3D Representation Methods}: SPA~\citep{zhu2025spa}, the prior SOTA of implicit 3D pretrained representation.
We adopt the three-layer MLP policy head for MetaWorld, diffusion-policy head~\citep{chi2023diffusion} for RLBench. For fairness, we keep the same training loss as GP3 for all models. For implicit 3D methods, we assess single-view (SPA-S and GP3-S) and multi-view configurations to investigate whether incorporating multiple viewpoints improves the quality of the learned geometric representations.

{\bfseries Training and Evaluation Details. } We use RoboVGGT as our implicit 3D foundation model. The model is finetuned for 40 epochs on a large-scale dataset of embodied robotic interactions collected in simulated and real-world scenarios, using multi-view RGB-D data and camera parameters to learn 3D-aware representations. The training runs on 8 H20 GPUs over 9 days.

For a fair comparison, all baselines follow a unified training and evaluation protocol, differing only in their visual modalities: 3D policy baselines additionally process 1024-point clouds generated with cropping and downsampling parameters as specified in their respective original papers, whereas other methods receive $224\times224$ RGB images. The robot state consists of the end-effector pose and joint angles, concatenated with the visual features.  
We use RoboVGGT as the implicit 3D foundation model. Following~\citep{jia2024lift3d}, we optimize using Adam $(\beta_1, \beta_2) = (0.95, 0.999)$ with a cosine annealing scheduler. The initial learning rate is set to $1\times10^{-3}$ for MetaWorld and $1\times10^{-4}$ for RLBench; both schedules include a linear warm-up over the first 10\% of training. Each method is trained for 100 epochs, and all trainings and evaluations run on H20 GPUs.

{\bfseries Quantitative Results. }For result of experiments Table~\ref{tab:combined_results}, across both MetaWorld and RLBench, GP3 achieves new state-of-the-art despite relying solely on RGB images.
On MetaWorld, it achieves an overall success rate of 86.7\%, surpassing the best prior implicit 3D representation (SPA) by 9.8\% and the best prior 3D policy (Lift3D) by 16.9\%.
On RLBench, GP3 attains 78.7\% with multi-view RGB, outperforming SPA by 26.7\% and Lift3D by 25.4\%. These results provide strong evidence that the implicit 3D representation learned by GP3 from RGB images consistently outperforms prior state-of-the-art methods that depend on explicit 3D geometric inputs.
Moreover, during multi-view training, the model acquires the capability to recover  point cloud from a single view as well. Owing to this, we obtain results that surpass those of other methods when inputting a single view.

\begin{table}[t]
\centering
\caption{Comparison of manipulation success rates on MetaWorld and RLBench benchmarks. Methods are grouped by feature and input type.}
\label{tab:combined_results}
\resizebox{\linewidth}{!}{%
\begin{tabular}{l c c | c c c c c | c}
\toprule
\multirow{2}{*}{\textbf{Method}} &
\multirow{2}{*}{\textbf{Feature Type}} &
\multirow{2}{*}{\textbf{Input Type}} &
\multicolumn{5}{c|}{\textbf{MetaWorld}} &
\multicolumn{1}{c}{\textbf{RLBench}} \\
\cmidrule(lr){4-8} \cmidrule(lr){9-9}
& & & \textbf{Easy} & \textbf{Med.} & \textbf{Hard} & \textbf{V.Hard} & \textbf{Mean} & \textbf{Mean} \\
\midrule
CLIP~\citep{radford2021learning_clip} & 2D Rep. & Single-view & 84.7 & 47.6 & 44.6 & 46.4 & 67.9 & 49.3 \\
R3M~\citep{nair2022rm} & 2D Rep. & Single-view & 80.4 & 49.5 & 55.3 & 38.4 & 66.4 & 66.0 \\
VC-1~\citep{majumdar2023we-vc1} & 2D Rep. & Single-view & 84.5 & 49.1 & 55.3 & 38.4 & 68.6 & 54.7 \\
\midrule
DP3~\citep{ze20243d_3dp} & 3D Rep. & Point Cloud & 79.3 & 34.2 & 44.0 & 29.6 & 60.1 & 57.3 \\
Lift3D~\citep{jia2024lift3d} & 3D Rep. & Point Cloud & 84.8 & 56.0 & 60.0 & 28.0 & 69.8 & 53.3 \\
\midrule
SPA-S~\citep{zhu2025spa} & 3D Rep. & Single-view & 86.0 & 53.8 & 48.0 & 46.4 & 70.4 &  60.0 \\
SPA & 3D Rep. & Multi-view & 91.0 & 62.2 & 58.6 & 58.4 & 77.5 & 52.0 \\
\midrule
\textbf{GP3-S (Ours)} & 3D Rep. & Single-view & 88.3 & 57.1 & 75.5 & 46.4 & 75.7 &  70.0 \\
\textbf{GP3 (Ours)} & 3D Rep. & Multi-view & \textbf{95.7} & \textbf{72.4} & \textbf{82.0} & \textbf{69.2} & \textbf{86.7} & \textbf{78.7} \\
\bottomrule
\end{tabular}%
}
\end{table}

\subsection{Real-World Experiment}~\label{section:E_2}

We performed a quantitative evaluation of our method in a real-world physical environment, built upon the open-source Mobile ALOHA platform~\citep{fu2024mobilealohalearningbimanual}. The experimental setup consists of two 7-DoF manipulator arms and three calibrated Intel RealSense D435i cameras. Two cameras are rigidly mounted on the robot wrists to capture actuator-centric views, while the third is fixed in a forward-facing position to provide a global scene perspective. This multi-view configuration supplies comprehensive visual information to the policy, which is essential for tasks requiring high precision and fine hand–eye coordination. We use a diffusion policy head for ALOHA, with training details kept the same as in the simulation experiments. We compare our method with ~\citep{zhu2025spa, chi2023diffusion}.

{\bfseries Dataset Collection. }We collected a teleoperated dataset for four common household-style tasks: Task1: Catching desktop paper ball and throw it away. Task2: Grasping the cloth and clean table.  Task3: Clearing items on desk. Task4: Retrieving a coffee from the desk.
For each task, we recorded 100 successful demonstration trajectories. Visual data from all three cameras was captured at 30 FPS, while the robot's proprioceptive state was recorded at the same FPS. To maintain synchronization with the visual inputs and improve efficiency during training, we down sampled the high-frequency action trajectories. Our policy model is trained to directly predict the robot's joint angles as its action output. To demonstrate the generalization capability of our spatial encoder, the tasks used for action training were excluded from the RoboVGGT training set.

{\bfseries Evaluation Protocol. }After training, the policy was deployed on the physical ALOHA robot for evaluation. While simulators can offer scalable and reproducible evaluation, real-world testing remains the definitive measure of performance. For each of the four tasks, we performed 20 independent trials and recorded the number of successful completions. Finally, we calculated the average success rate for each task to quantitatively assess the real-world performance and robustness of our method.

\begin{figure}[!htb]
    \centering
    \subfloat[Original Experiment: Grab and Throw Paper Ball.] {
        \includegraphics[width=0.32\columnwidth]{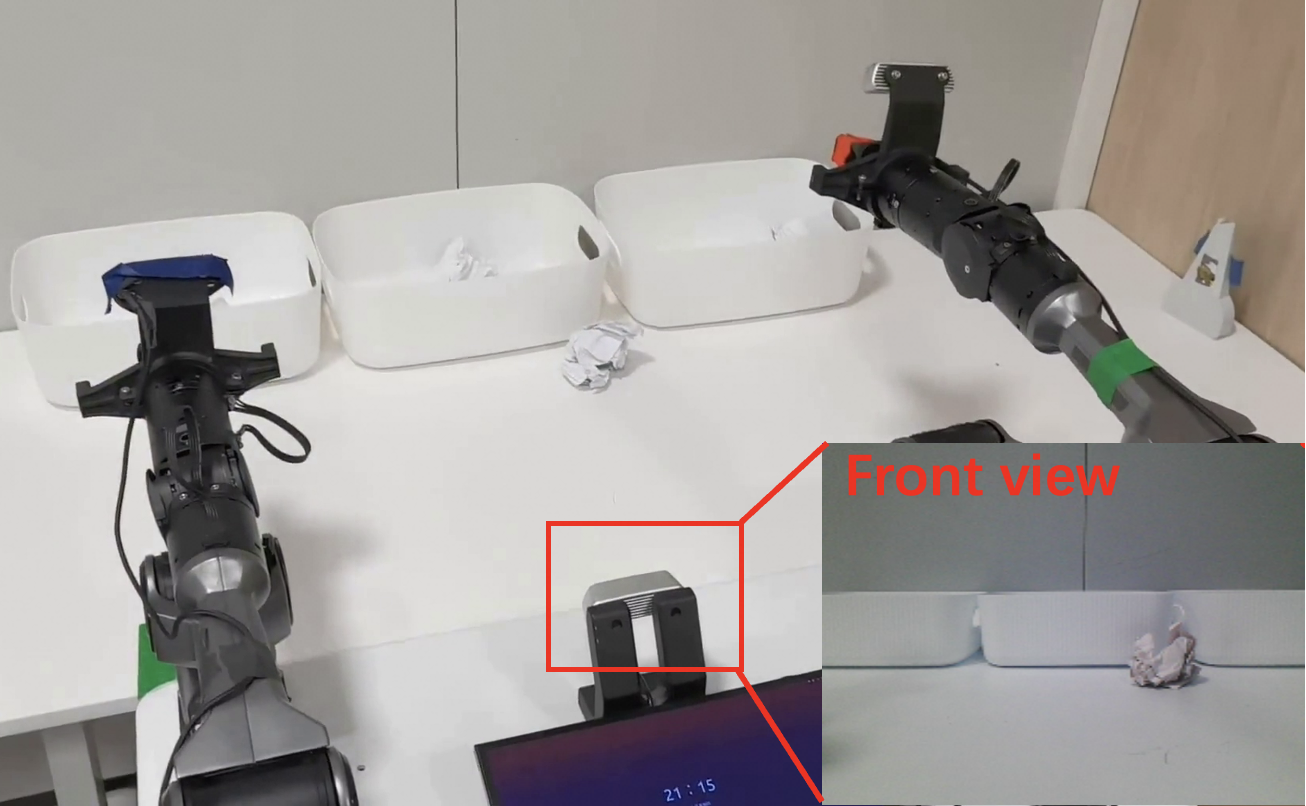}
    }
    \hspace{0.1\columnwidth}  
    \subfloat[Comparative Experiment: Replacing Crumpled Paper with Printed Flat Paper.] {
        \includegraphics[width=0.32\columnwidth]{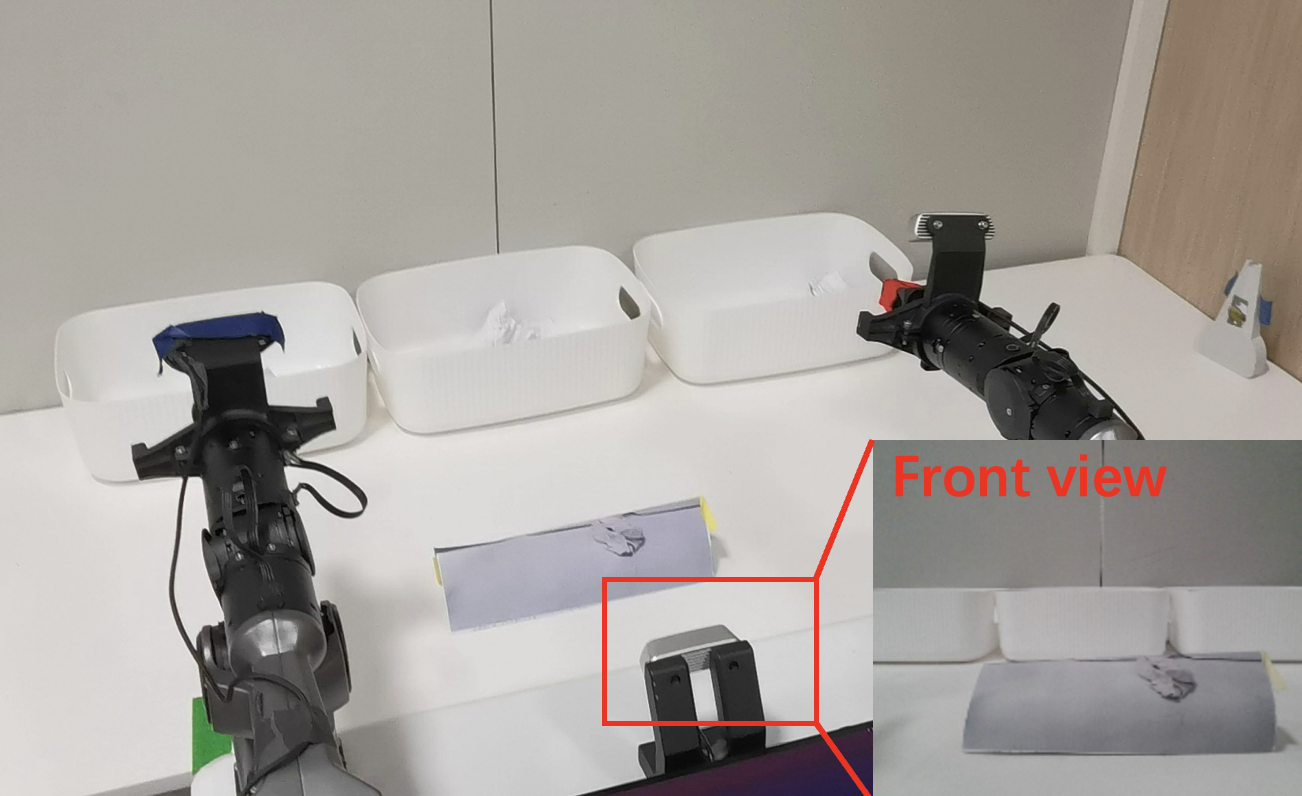}
    }
    \caption{Experimental setup for 3D perception validation, in which only the multi-view input GP3 is not deceived by the printed flat paper.}
    \label{fig:3d awareness setup}
\end{figure}

{\bfseries Quantitative Results. }As shown in Table~\ref{tab:real_result}, GP3 with multi-view input achieves state-of-the-art performance on all evaluated real-world tasks. Unlike other methods that benefit little or even perform worse in multi perspective situations, our method shows significant optimization as the camera increases. To further verify the superiority of the proposed method in 3D perception, we conduct a comparative experiment, with the setup illustrated in Figure~\ref{fig:3d awareness setup} and detailed configurations provided in Figure~\ref{fig:gp3_3d_understanding}. While Diffusion Policy and SPA with multi-view input, together with GP3 with single-view input, are vulnerable to deception, GP3 with multi-view input maintains robustness and outperforms the baseline in complex spatial understanding tasks.

\begin{table}[!htb]
\centering
\caption{Comparison of manipulation success rates on the real-world experiments.}
\label{tab:real_result}
\small 
\setlength{\tabcolsep}{4pt} 
\begin{tabular*}{\linewidth}{@{\extracolsep{\fill}}l c c |
 c c c c}
\toprule
\textbf{Method} & \textbf{Feature Type} & \textbf{View} & \textbf{Task 1} & \textbf{Task 2} & \textbf{Task 3} & \textbf{Task 4} \\
\midrule
Diffusion Policy-S~\citep{chi2023diffusion} & 2D Rep. & 1 & 8/20 & 5/20 & 3/20 & 1/20 \\
Diffusion Policy~\citep{chi2023diffusion} & 2D Rep. & 3 & 5/20 & 7/20 & 1/20 & 0/20 \\
SPA-S~\citep{zhu2025spa} & 3D Rep. & 1 & 3/20 & 6/20 & 3/20 & 3/20 \\
SPA~\citep{zhu2025spa} & 3D Rep. & 3 & 7/20 & 7/20 & 5/20 & 4/20 \\
\midrule
\textbf{GP3-S (Ours)} & 3D Rep. & 1 & 14/20 & 13/20 & 11/20 & 13/20 \\
\textbf{GP3 (Ours)} & 3D Rep. & 3 & \textbf{18/20} & \textbf{19/20} & \textbf{15/20} & \textbf{17/20} \\
\bottomrule
\end{tabular*}
\end{table}

\subsection{Ablation Study}~\label{section:E_3}

To systematically assess the contributions of each core component, the fine-tuning of spatial encoder for robot scene reconstruction, and the language guidance mechanism within our framework, we conducted an ablation study on 15 tasks from the MetaWorld benchmark. These tasks are categorized as follows: easy tasks include button-press, drawer-open, reach, handle-pull, peg-unplug-side, lever-pull, and dial-turn; medium tasks include hammer, sweep-into, bin-picking, push-wall, and box-close; hard tasks include assembly, hand-insert, and shelf-place. For Quantitative results are summarized in Table~\ref{tab:ablation}, and the key findings are as follows:

\subsubsection{Impact of Fine-tuning}
The baseline model (without fine-tuning) achieves a 56.0\% mean success rate with single-view inputs. After fine-tuning on robot-centric data, the single-view success rate improves to 70.9\%, highlighting the importance of adaptation to robot-specific scenarios. Similarly, the 2-view and 4-view configurations show increases of 12.3\% and 15.5\%, respectively, after fine-tuning.

\subsubsection{Multi-view Inputs}
Compared to its single-view baseline (56.0\%), the model’s performance improves to 66.1\% with two views. However, increasing to four views unexpectedly reduces performance to 61.8\%. Even after fine-tuning on robot-centric datasets (w/ FT), the success rate declines from 78.4\% (two views) to 77.3\% (four views). This suggests that once the number of views exceeds an optimal threshold, standard geometric models without specific view-handling mechanisms may introduce redundancy and noise, which in turn leads to degraded performance.

\subsubsection{Conventional FiLM-based Feature Modulation}
The conventional FiLM implementation (w/ FT+F), when combined with the fine-tuned encoder, shows a positive impact, generally improving upon the w/ FT baseline: With 4 views, w/ FT+F achieves an 80.3\% mean success rate, a 3.0\% improvement over w/ FT (4 views, 77.3\%), and 1.3\% improvement over w/ FT+F (2 views, 79.0\%). This indicates that FiLM, by incorporating linguistic features, helps aggregate and modulate features from different views.

\subsubsection{G-FiLM Architecture}
With 4 views, w/ FT+GF leads with an 83.9\% mean success rate, representing a 3.6\% improvement over w/ FT+F (4 views, 80.3\%), and a 2.3\% improvement over w/ FT+GF (2 views, 81.6\%). 
The multi-view improvement surpasses that of using conventional FiLM, demonstrating that G-FiLM better handles noise and redundancy in multi-view inputs compared to FiLM. This validates the advantage of selectively fusing geometric and linguistic features, enabling the incorporation of semantic cues into the multi-view fusion module while preserving the geometric model’s inherent capability for scene feature extraction. Such a design is particularly critical for robotic manipulation in complex environments with multi-view observations.

\begin{table}[ht]
\centering
\caption{Ablation study on a 15-task MetaWorld subset. FT means using fine-tuned encoder on robot dataset, F means using FiLM, and GF means using G-FiLM.}
\label{tab:ablation}
\small 
\setlength{\tabcolsep}{0pt} 
\begin{tabular*}{\linewidth}{@{\extracolsep{\fill}}l c c c c c}
\toprule
\textbf{Method} & \textbf{Views} & \textbf{Easy} & \textbf{Medium} & \textbf{Hard} & \textbf{Mean S.R.} \\
\midrule
\multirow{3}{*}{\textbf{Baseline}} 
& 1 & 63.4 & 55.2 & 40.0 & 56.0 \\
& 2 & 70.2 & 69.6 & 50.7 & 66.1 \\
& 4 & 74.3 & 57.6 & 40.0 & 61.8 \\
\cmidrule(lr){2-6}
\multirow{3}{*}{\textbf{w/ FT}} 
& 1 & 80.0 & 64.8 & 60.0 & 70.9 \\
& 2 & 83.4 & 74.4 & 73.4 & 78.4 \\
& 4 & 82.3 & 76.0 & 68.0 & 77.3 \\
\cmidrule(lr){2-6}
\multirow{3}{*}{\textbf{w/ FT+F}} 
& 1 & 79.4 & 75.2 & 57.3 & 73.6 \\
& 2 & 85.0 & 73.6 & 72.0 & 79.0 \\
& 4 & 86.3 & 74.4 & 76.0 & 80.3 \\
\cmidrule(lr){2-6}
\multirow{3}{*}{\textbf{w/ FT+GF}} 
& 1 & 78.8 & 80.8 & 61.3 & 76.0 \\
& 2 & 81.7 & 83.2 & \textbf{78.6} & 81.6 \\
& 4 & \textbf{86.9} & \textbf{85.7} & 69.3 & \textbf{83.9} \\
\bottomrule
\end{tabular*}
\end{table}

\subsection{Point Map Estimation}~\label{section:E_4}
We further compare the accuracy of our predicted point clouds with that of the original VGGT on two unseen RoboTwin tasks, whose scenes are significantly different from those in the training dataset. We generate 20 demonstrations per task and randomly sample 10 frames for each demonstration. Predicted point clouds are aligned to the ground truth using the Umeyama~\citep{88573_Umeyama} algorithm. Following~\citep{10378090_PoseDiffusion}, we report \textit{Accuracy}, \textit{Completeness}, and \textit{Overall} (Chamfer distance) for point map estimation. To ensure the reliability of the evaluation, we filter out views with an overlap of less than 0.35 with the main view during validation. This threshold is chosen to exclude low-quality or highly dissimilar views that could negatively affect the accuracy of the comparison.
As shown in Table~\ref{tab:vs_vggt}, fine-tuning improves the geometry model’s performance in robotic scenes.

\begin{table}[ht]
\centering
\caption{Dense MVS estimation on an unseen RoboTwin task. Lower values are better ($\downarrow$).}
\label{tab:vs_vggt}
\setlength{\tabcolsep}{0pt}
\begin{tabular*}{\linewidth}{@{\extracolsep{\fill}}l r r r}
\toprule
\textbf{Method} & \textbf{Acc.} $\downarrow$ & \textbf{Comp.} $\downarrow$ & \textbf{Overall} $\downarrow$ \\
\midrule
Original VGGT & 4.756 & 0.923 & 2.840 \\
RoboVGGT      & 1.142 & 0.472 & 0.807 \\
\bottomrule
\end{tabular*}
\end{table}

\section{Conclusion and Future Work}
\label{sec:conclusion}

We propose GP3, a 3D visual geometry-aware policy learning framework that enables sensor-agnostic visuomotor control using only multi-view RGB inputs. By fine-tuning a large-scale 3D reconstruction model, GP3 achieves robust spatial reasoning without depth sensors or explicit 3D data. Our experiments demonstrate state-of-the-art performance on MetaWorld and RLBench benchmarks, outperforming both 2D-based and 3D-based methods. The framework successfully transfers to real-world robotic systems with minimal overhead, validating its practical viability.
The success of GP3 establishes that robust 3D spatial reasoning for manipulation can be achieved end-to-end from RGB inputs, making it a lightweight, scalable, and highly generalizable solution. 

Currently, GP3 is limited to interpreting multiple perspectives simultaneously, future work will focus on extending the framework to handle long-horizon, dynamic tasks and exploring its application to more complex manipulation skills.

\bibliographystyle{assets/plainnat}
\bibliography{paper}

\clearpage
\newpage
\beginappendix

\section{More Details about RoboVGGT Data Curation}
Here we detail the data curation procedure used to fine-tune RoboVGGT for improved efficiency and embodiment-scene generalization.

\subsection{Simulated Benchmarks}
\begin{figure}[!htb]
    \centering
    \begin{subfigure}[t]{0.22\columnwidth}
        \centering
        \includegraphics[width=\linewidth]{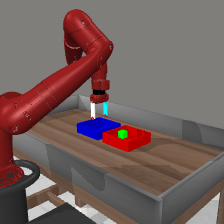}
        \caption{Corner}
        \label{fig:sub1}
    \end{subfigure}
    \hfill
    \begin{subfigure}[t]{0.22\columnwidth}
        \centering
        \includegraphics[width=\linewidth]{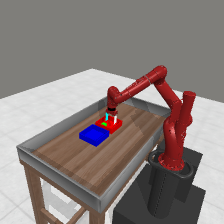}
        \caption{Corner2}
        \label{fig:sub2}
    \end{subfigure}
    \hfill
    \begin{subfigure}[t]{0.22\columnwidth}
        \centering
        \includegraphics[width=\linewidth]{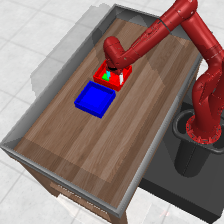}
        \caption{Corner3}
        \label{fig:sub3}
    \end{subfigure}
    \hfill
    \begin{subfigure}[t]{0.22\columnwidth}
        \centering
        \includegraphics[width=\linewidth]{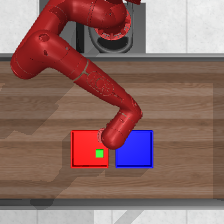}
        \caption{Topview}
        \label{fig:sub4}
    \end{subfigure}
    \caption{Multi-view MetaWorld visualization showing the four camera viewpoints used for data collection.}
    \label{fig:multi-view_metaworld}
\end{figure}

{\bfseries MetaWorld~\citep{yu2020meta}. }This benchmark consists of a series of tasks in which an agent controls a single-arm Sawyer robot to manipulate objects in a tabletop environment. We selected all 50 diverse manipulation tasks in MetaWorld, including pushing, pulling, placing, and tool use, and others. To support rich visual perception and multi-view learning, we collected data from four fixed camera viewpoints: \emph{corner}, \emph{corner2}, \emph{corner3}, and \emph{topview}, which provide complementary perspectives of the scene. Figure~\ref{fig:multi-view_metaworld} demonstrates the multi-view observation setup used in this benchmark. 

\begin{figure}[!htb]
    \centering
    \begin{subfigure}[t]{0.18\columnwidth}
        \centering
        \includegraphics[width=\linewidth]{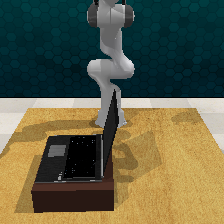}
        \caption{Front}
        \label{fig:sub1}
    \end{subfigure}
    \hfill
    \begin{subfigure}[t]{0.18\columnwidth}
        \centering
        \includegraphics[width=\linewidth]{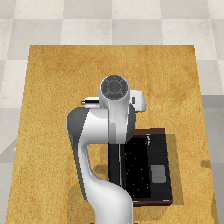}
        \caption{Overhead}
        \label{fig:sub2}
    \end{subfigure}
    \hfill
    \begin{subfigure}[t]{0.18\columnwidth}
        \centering
        \includegraphics[width=\linewidth]{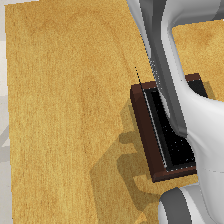}
        \caption{Left Shoulder}
        \label{fig:sub3}
    \end{subfigure}
    \hfill
    \begin{subfigure}[t]{0.18\columnwidth}
        \centering
        \includegraphics[width=\linewidth]{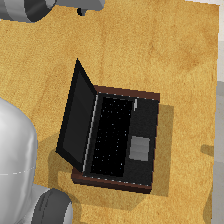}
        \caption{Right Shoulder}
        \label{fig:sub4}
    \end{subfigure}
    \hfill
    \begin{subfigure}[t]{0.18\columnwidth}
        \centering
        \includegraphics[width=\linewidth]{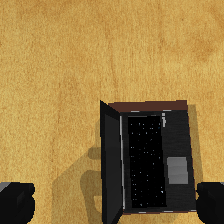}
        \caption{Wrist}
        \label{fig:sub5}
    \end{subfigure}
    \caption{Multi-view RLBench visualization showing the five camera viewpoints used for data collection.}
    \label{fig:multi-view_rlbench}
\end{figure}

{\bfseries RLBench~\citep{james2020rlbench}. }This benchmark consists of a series of tasks performed by a Franka Emika Panda robotic arm in a simulated tabletop environment. We selected 14 diverse manipulation tasks from RLBench~\citep{james2020rlbench}. To support multi-view representation learning, we collected synchronized RGB data from five standard camera viewpoints: \emph{front}, \emph{overhead}, \emph{left\_shoulder}, \emph{right\_shoulder}, and \emph{wrist}. These views provide complementary visual information, including scene-level context and egocentric observation. Figure~\ref{fig:multi-view_rlbench} illustrates the multi-view observation setup used in this benchmark.

\begin{figure}[H]
\centering
\captionsetup[subfigure]{justification=centering}
\setcounter{subfigure}{0}

\begin{subfigure}[t]{0.24\linewidth}
    \centering
    \includegraphics[width=\linewidth]{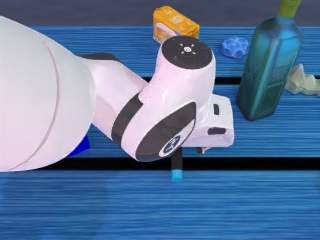}
    \caption{Head}
    \label{fig:franka_head}
\end{subfigure}
\hfill
\begin{subfigure}[t]{0.24\linewidth}
    \centering
    \includegraphics[width=\linewidth]{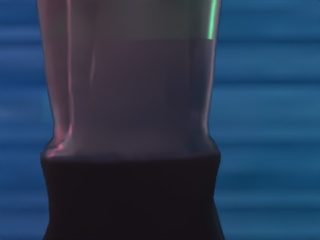}
    \caption{Left}
    \label{fig:franka_left}
\end{subfigure}
\hfill
\begin{subfigure}[t]{0.24\linewidth}
    \centering
    \includegraphics[width=\linewidth]{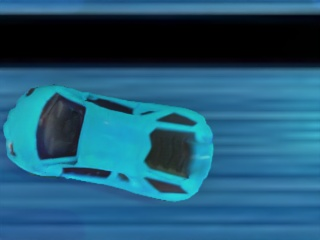}
    \caption{Right}
    \label{fig:franka_right}
\end{subfigure}

\vspace{2mm} 

\begin{subfigure}[t]{0.24\linewidth}
    \centering
    \includegraphics[width=\linewidth]{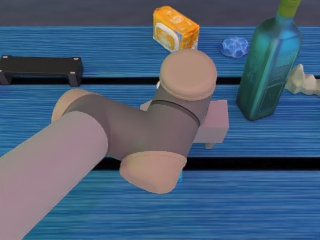}
    \caption{Head}
    \label{fig:ur5_head}
\end{subfigure}
\hfill
\begin{subfigure}[t]{0.24\linewidth}
    \centering
    \includegraphics[width=\linewidth]{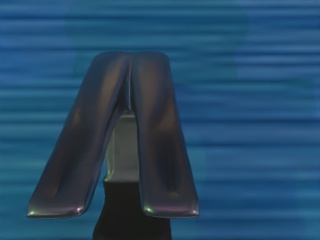}
    \caption{Left}
    \label{fig:ur5_left}
\end{subfigure}
\hfill
\begin{subfigure}[t]{0.24\linewidth}
    \centering
    \includegraphics[width=\linewidth]{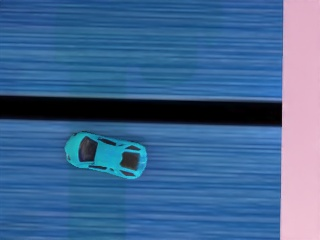}
    \caption{Right}
    \label{fig:ur5_right}
\end{subfigure}

\vspace{2mm}

\begin{subfigure}[t]{0.24\linewidth}
    \centering
    \includegraphics[width=\linewidth]{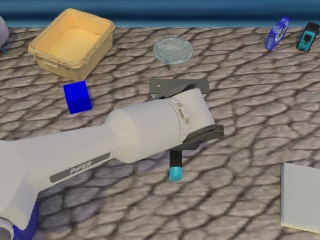}
    \caption{Head}
    \label{fig:piper_head}
\end{subfigure}
\hfill
\begin{subfigure}[t]{0.24\linewidth}
    \centering
    \includegraphics[width=\linewidth]{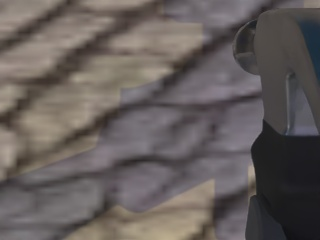}
    \caption{Left}
    \label{fig:piper_left}
\end{subfigure}
\hfill
\begin{subfigure}[t]{0.24\linewidth}
    \centering
    \includegraphics[width=\linewidth]{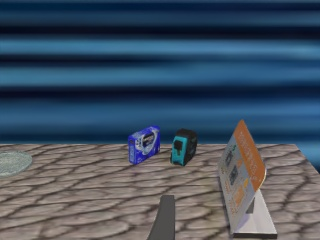}
    \caption{Right}
    \label{fig:piper_right}
\end{subfigure}

\vspace{2mm}

\begin{subfigure}[t]{0.24\linewidth}
    \centering
    \includegraphics[width=\linewidth]{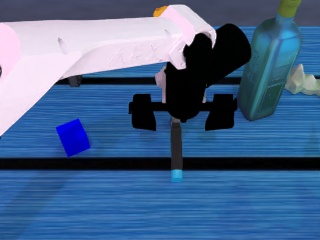}
    \caption{Head}
    \label{fig:arx_head}
\end{subfigure}
\hfill
\begin{subfigure}[t]{0.24\linewidth}
    \centering
    \includegraphics[width=\linewidth]{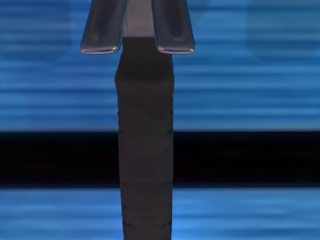}
    \caption{Left}
    \label{fig:arx_left}
\end{subfigure}
\hfill
\begin{subfigure}[t]{0.24\linewidth}
    \centering
    \includegraphics[width=\linewidth]{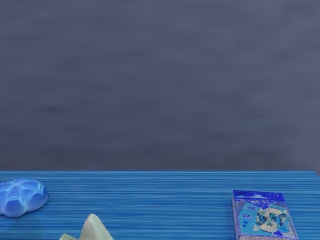}
    \caption{Right}
    \label{fig:arx_right}
\end{subfigure}

\vspace{3mm}

\begin{subfigure}[t]{0.19\linewidth}
    \centering
    \includegraphics[width=\linewidth]{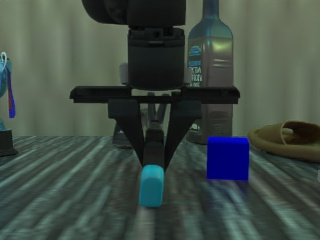}
    \caption{Front}
    \label{fig:rand_front}
\end{subfigure}
\hfill
\begin{subfigure}[t]{0.19\linewidth}
    \centering
    \includegraphics[width=\linewidth]{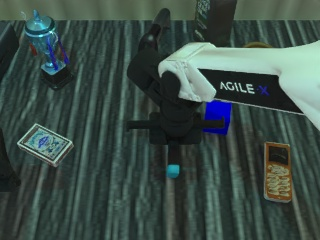}
    \caption{Head}
    \label{fig:rand_head}
\end{subfigure}
\hfill
\begin{subfigure}[t]{0.19\linewidth}
    \centering
    \includegraphics[width=\linewidth]{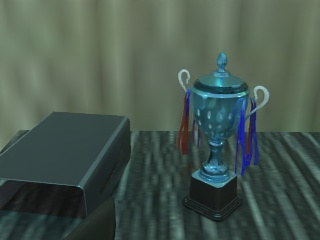}
    \caption{Left}
    \label{fig:rand_left}
\end{subfigure}
\hfill
\begin{subfigure}[t]{0.19\linewidth}
    \centering
    \includegraphics[width=\linewidth]{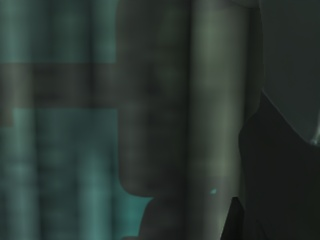}
    \caption{Right}
    \label{fig:rand_right}
\end{subfigure}

\caption{Multi-view observations across five robot configurations in the RoboTwins dataset. Each group shows the camera views used for that robot, with the robot name displayed below as a section header. The robot embodiments are arranged from top to bottom in the following order: Arx-X5, Franka, Piper, UR5-WSG, and Agile-X.}
\label{fig:multi-view_robotwins}
\end{figure}

{\bfseries RoboTwins~\citep{mu2025robotwin}. }This benchmark includes five diverse robot configurations: the \textbf{Franka Emika Panda}, the \textbf{UR5 with a WSG-50 gripper}, \textbf{Piper}, \textbf{ARX-X5}, and \textbf{Agile-X}. The variety in kinematics, appearance, and dynamics across these robots enhances the generalization capability of RoboVGGT by encouraging learning of robot-agnostic visual representations. Except for the Agile-X variant, which includes four views (\emph{front}, \emph{head}, \emph{left}, \emph{right}), all other configurations provide three views: \emph{head}, \emph{left}, and \emph{right}. Figure~\ref{fig:multi-view_robotwins} demonstrates the multi-view observation setup used in this benchmark. 

We collected 36 tasks including: Adjust Bottle, Beat Block With Hammer, Blocks Ranking By Color, Blocks Ranking By Size, Click Bell, Grab Roller, Hand Over Block, Lift Pot, Move Can To Pot, Move Pill Bottle To Pad, Move Playing Card Away, Move Stapler To Pad, Pick Diverse Bottles, Pick Two Bottles, Place A Into B Left, Place A Into B Right, Place Bread In Basket, Place Bread In Skillet, Place Burger And Fries, Place Can In Basket, Place Cans In Plastic Box, Place Container On Plate, Place Two Shoes, Place Empty Cup, Place Fan, Place Mouse Pad, Place Object In Basket, Place Object On Scale, Place Object On Stand, Place Phone On Stand, Place Shoe, Press Stapler, Rotate QR Code, Scan Object, Stack Three Blocks, Stack Two Blocks.

\subsection{Real-World Benchmarks}
\begin{figure}[!htb]
    \centering
    \begin{subfigure}[t]{0.24\columnwidth}
        \centering
        \includegraphics[width=\linewidth]{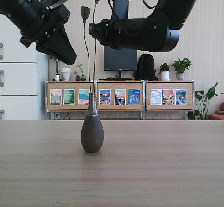}
        \caption{Front}
        \label{fig:sub1}
    \end{subfigure}
    \hfill
    \begin{subfigure}[t]{0.24\columnwidth}
        \centering
        \includegraphics[width=\linewidth]{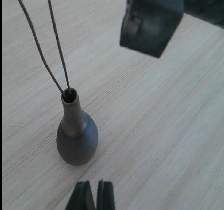}
        \caption{Left}
        \label{fig:sub2}
    \end{subfigure}
    \hfill
    \begin{subfigure}[t]{0.24\columnwidth}
        \centering
        \includegraphics[width=\linewidth]{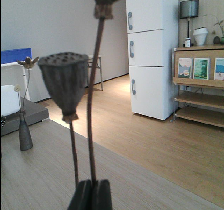}
        \caption{Right}
        \label{fig:sub3}
    \end{subfigure}
    \caption{Multi-view Aloha visualization showing the three camera viewpoints used for data collection.}
    \label{fig:multi-aloha}
\end{figure}

In our real-world experiments, we use the open-source Mobile ALOHA platform~\citep{fu2024mobilealohalearningbimanual}, which is equipped with three RealSense D435i depth cameras: one front-facing camera and one mounted on each robotic arm. The cameras' intrinsic and extrinsic parameters are calibrated to obtain accurate ground-truth alignments. We further train RoboVGGT on several daily manipulation tasks, including: Clean the Table, Press Button, Flower Arrangement, Load Sundries into Bag, Pour Water, Pick Up Doll, and Make Coffee. Figure~\ref{fig:multi-aloha} demonstrates the multi-view observation setup used in this benchmark. 

\section{More Simulation Results}

In this section, we provide additional results on the multi-task setup described in Section~\ref{section:E_1}. We also present detailed performance of each task under different experimental settings to further analyze the effectiveness and generalization of the proposed method.

The experimental setup follows the same protocol as described in Section~\ref{section:E_1}, with the exception that GP3 and all baselines are trained and evaluated on a multi-task setting instead of a single-task. Specifically, for MetaWorld, we train and evaluate on all 50 tasks in the benchmark, and for RLBench, we use 6 tasks following the setup of LIFT3D~\citep{jia2024lift3d}.

As shown in Table~\ref{tab:combined_results}, GP3 also achieves the state-of-the-art performance on the multi-task setting. The overall success rates of GP3 reach 83.1\% on MetaWorld and 83.3\% on RLBench, significantly outperforming the best baseline methods in each respective task. Specifically, on MetaWorld, GP3 surpasses the top-performing baseline SPA by 9.8\%, and on RLBench, it exceeds R3M by 24.6\%. These results further demonstrate the superior generalization capability and performance of GP3 across both single-task and multi-task settings.

\begin{table}[t]
\centering
\caption{Multi-Task Performance Comparison on MetaWorld and RLBench Benchmarks.}
\label{tab:combined_results2}
\resizebox{\linewidth}{!}{%
\begin{tabular}{l c c | c c c c c | c}
\toprule
\multirow{2}{*}{\textbf{Method}} &
\multirow{2}{*}{\textbf{Feature Type}} &
\multirow{2}{*}{\textbf{Input Type}} &
\multicolumn{5}{c|}{\textbf{MetaWorld}} &
\multicolumn{1}{c}{\textbf{RLBench}} \\
\cmidrule(lr){4-8} \cmidrule(lr){9-9}
& & & \textbf{Easy} & \textbf{Med.} & \textbf{Hard} & \textbf{V.Hard} & \textbf{Mean} & \textbf{Mean} \\
\midrule
CLIP~\citep{radford2021learning_clip} & 2D Rep. & Single-view & 80.6 & 48.7 & 38.7 & 37.6 & 64.2 & 53.3 \\
R3M~\citep{nair2022rm} & 2D Rep. & Single-view & 83.8 & 52.4 & 40.0 & 44.0 & 67.7 & 58.7 \\
VC-1~\citep{majumdar2023we-vc1} & 2D Rep. & Single-view & 82.1 & 50.9 & 38.6 & 32.0 & 65.0 & 58.7 \\
\midrule
DP3~\citep{ze20243d_3dp} & 3D Rep. & Point Cloud & 52.9 & 13.8 & 10.0 & 15.2 & 35.4 & 50.7 \\
Lift3D~\citep{jia2024lift3d} & 3D Rep. & Point Cloud & 71.9 & 31.3 & 26.0 & 41.6 & 54.4 & 59.0 \\
\midrule
SPA-S~\citep{zhu2025spa} & 3D Rep. & Single-view & 81.6 & 52.4 & 41.3 & 43.2 & 66.5 & 56.0 \\
SPA & 3D Rep. & Multi-view & 86.6 & 60.7 & 52.0 & 52.0 & 73.3 & 50.0 \\
\midrule
\textbf{GP3-S (Ours)} & 3D Rep. & Single-view & 81.3 & 78.5  & 76.0 & 55.2 & 77.4 & 80.0 \\
\textbf{GP3 (Ours)} & 3D Rep. & Multi-view & \textbf{85.9} & \textbf{77.5} & \textbf{90.0} & \textbf{72.0} & \textbf{83.1} & \textbf{83.3} \\
\bottomrule
\end{tabular}%
}
\end{table}


\renewcommand{\arraystretch}{1.1}  
\setlength{\tabcolsep}{0pt}        

\begin{table*}[htbp]
\centering
\caption{All single-task results on RLBench.}
\begin{adjustbox}{width=\linewidth, totalheight=\textheight, keepaspectratio, scale=0.98}
\scriptsize 

\newcommand{\methodcolwidth}{1.7em}   
\newcommand{\taskcolwidth}{7.0em}    

\begin{tabular}{
    >{\raggedright\arraybackslash\hspace{0pt}\scriptsize}p{\taskcolwidth}  
    *{9}{ 
        |>{\centering\arraybackslash}p{\methodcolwidth}
         >{\centering\arraybackslash}p{\methodcolwidth}
         >{\centering\arraybackslash}p{\methodcolwidth}
    }
}
\toprule
\multirow{2}{*}{\textbf{Method}} &
\multicolumn{3}{c|}{\multirow{2}{*}{\textbf{CLIP}}} &
\multicolumn{3}{c|}{\multirow{2}{*}{\textbf{R3M}}} &
\multicolumn{3}{c|}{\multirow{2}{*}{\textbf{VC-1}}} &
\multicolumn{3}{c|}{\multirow{2}{*}{\textbf{DP3}}} &
\multicolumn{3}{c|}{\multirow{2}{*}{\textbf{Lift3d}}} &
\multicolumn{3}{c|}{\multirow{2}{*}{\textbf{SPA-S}}} &
\multicolumn{3}{c|}{\multirow{2}{*}{\textbf{SPA}}} &
\multicolumn{3}{c|}{\multirow{2}{*}{\textbf{GP3-S}}} &
\multicolumn{3}{c}{\multirow{2}{*}{\textbf{GP3}}} \\

& & & & & & & & & & & & & & & & & & & & & & & & & & & \\
\midrule
\textbf{CloseBox}       & \multicolumn{3}{c|}{76} & \multicolumn{3}{c|}{100} & \multicolumn{3}{c|}{92} & \multicolumn{3}{c|}{92} & \multicolumn{3}{c|}{96} & \multicolumn{3}{c|}{96} & \multicolumn{3}{c|}{100} &\multicolumn{3}{c|}{96} & \multicolumn{3}{c}{96} \\
\textbf{PutRubbishInBin}   & \multicolumn{3}{c|}{20} & \multicolumn{3}{c|}{52} & \multicolumn{3}{c|}{8} & \multicolumn{3}{c|}{28} & \multicolumn{3}{c|}{8} & \multicolumn{3}{c|}{12} & \multicolumn{3}{c|}{0} &\multicolumn{3}{c|}{52} & \multicolumn{3}{c}{68} \\
\textbf{CloseLaptopLid}          & \multicolumn{3}{c|}{96} & \multicolumn{3}{c|}{96} & \multicolumn{3}{c|}{88} & \multicolumn{3}{c|}{84} & \multicolumn{3}{c|}{80} & \multicolumn{3}{c|}{96} & \multicolumn{3}{c|}{60} &\multicolumn{3}{c|}{100} & \multicolumn{3}{c}{96} \\
\textbf{WaterPlants}             &  \multicolumn{3}{c|}{12} & \multicolumn{3}{c|}{16} & \multicolumn{3}{c|}{12} & \multicolumn{3}{c|}{12} & \multicolumn{3}{c|}{4} & \multicolumn{3}{c|}{16} & \multicolumn{3}{c|}{20} &\multicolumn{3}{c|}{36} & \multicolumn{3}{c}{44} \\
\textbf{UnplugCharger}                &  \multicolumn{3}{c|}{16} & \multicolumn{3}{c|}{32} & \multicolumn{3}{c|}{32} & \multicolumn{3}{c|}{52} & \multicolumn{3}{c|}{48} & \multicolumn{3}{c|}{40} & \multicolumn{3}{c|}{36} &\multicolumn{3}{c|}{36} & \multicolumn{3}{c}{72} \\
\textbf{ToiletSeatDown}                 & \multicolumn{3}{c|}{76} & \multicolumn{3}{c|}{100} & \multicolumn{3}{c|}{96} & \multicolumn{3}{c|}{76} & \multicolumn{3}{c|}{84} & \multicolumn{3}{c|}{100} & \multicolumn{3}{c|}{96} &\multicolumn{3}{c|}{100} & \multicolumn{3}{c}{96} \\
\bottomrule
\end{tabular}
\end{adjustbox}
\end{table*}

\begin{table*}[htbp]
\centering
\caption{All single-task results on Meta-World.}
\begin{adjustbox}{width=\linewidth, totalheight=\textheight, keepaspectratio, scale=0.98}
\scriptsize 

\newcommand{\methodcolwidth}{1.7em}   
\newcommand{\taskcolwidth}{10.0em}    

\begin{tabular}{
    >{\raggedright\arraybackslash\hspace{0pt}\scriptsize}p{\taskcolwidth}  
    *{9}{ 
        |>{\centering\arraybackslash}p{\methodcolwidth}
         >{\centering\arraybackslash}p{\methodcolwidth}
         >{\centering\arraybackslash}p{\methodcolwidth}
    }
}
\toprule
\multirow{2}{*}{\textbf{Method}} &
\multicolumn{3}{c|}{\multirow{2}{*}{\textbf{CLIP}}} &
\multicolumn{3}{c|}{\multirow{2}{*}{\textbf{R3M}}} &
\multicolumn{3}{c|}{\multirow{2}{*}{\textbf{VC-1}}} &
\multicolumn{3}{c|}{\multirow{2}{*}{\textbf{DP3}}} &
\multicolumn{3}{c|}{\multirow{2}{*}{\textbf{Lift3d}}} &
\multicolumn{3}{c|}{\multirow{2}{*}{\textbf{SPA-S}}} &
\multicolumn{3}{c|}{\multirow{2}{*}{\textbf{SPA}}} &
\multicolumn{3}{c|}{\multirow{2}{*}{\textbf{GP3-S}}} &
\multicolumn{3}{c}{\multirow{2}{*}{\textbf{GP3}}} \\

& & & & & & & & & & & & & & & & & & & & & & & & & & & \\
\midrule
\textbf{ButtonPress}              & \multicolumn{3}{c|}{100} & \multicolumn{3}{c|}{100}  & \multicolumn{3}{c|}{100} & \multicolumn{3}{c|}{100} & \multicolumn{3}{c|}{100} & \multicolumn{3}{c|}{100} & \multicolumn{3}{c|}{100} & \multicolumn{3}{c|}{100} & \multicolumn{3}{c}{100} \\
\textbf{ButtonPressTopdown}       & \multicolumn{3}{c|}{100} & \multicolumn{3}{c|}{100}  & \multicolumn{3}{c|}{100} & \multicolumn{3}{c|}{96} & \multicolumn{3}{c|}{100} & \multicolumn{3}{c|}{100} & \multicolumn{3}{c|}{100} & \multicolumn{3}{c|}{100} & \multicolumn{3}{c}{100} \\
\textbf{ButtonPressTopdownWall}   & \multicolumn{3}{c|}{100} & \multicolumn{3}{c|}{96}   & \multicolumn{3}{c|}{96}  & \multicolumn{3}{c|}{100} & \multicolumn{3}{c|}{100} & \multicolumn{3}{c|}{100} & \multicolumn{3}{c|}{100} & \multicolumn{3}{c|}{100} & \multicolumn{3}{c}{100} \\
\textbf{ButtonPressWall}          & \multicolumn{3}{c|}{100} & \multicolumn{3}{c|}{100}  & \multicolumn{3}{c|}{100} & \multicolumn{3}{c|}{100} & \multicolumn{3}{c|}{100} & \multicolumn{3}{c|}{100} & \multicolumn{3}{c|}{100} & \multicolumn{3}{c|}{100} & \multicolumn{3}{c}{100} \\
\textbf{CoffeeButton}             & \multicolumn{3}{c|}{100} & \multicolumn{3}{c|}{100}  & \multicolumn{3}{c|}{100} & \multicolumn{3}{c|}{100} & \multicolumn{3}{c|}{100} & \multicolumn{3}{c|}{100} & \multicolumn{3}{c|}{100} & \multicolumn{3}{c|}{100} & \multicolumn{3}{c}{100} \\
\textbf{DialTurn}                 & \multicolumn{3}{c|}{64}  & \multicolumn{3}{c|}{80}   & \multicolumn{3}{c|}{80}  & \multicolumn{3}{c|}{100} & \multicolumn{3}{c|}{96} & \multicolumn{3}{c|}{64} & \multicolumn{3}{c|}{84} & \multicolumn{3}{c|}{88} & \multicolumn{3}{c}{96} \\
\textbf{DoorClose}                & \multicolumn{3}{c|}{100} & \multicolumn{3}{c|}{100}  & \multicolumn{3}{c|}{100} & \multicolumn{3}{c|}{100} & \multicolumn{3}{c|}{100} & \multicolumn{3}{c|}{100} & \multicolumn{3}{c|}{100} & \multicolumn{3}{c|}{100} & \multicolumn{3}{c}{100} \\
\textbf{DoorLock}                 & \multicolumn{3}{c|}{48}  & \multicolumn{3}{c|}{0}    & \multicolumn{3}{c|}{0}   & \multicolumn{3}{c|}{84} & \multicolumn{3}{c|}{96} & \multicolumn{3}{c|}{44} & \multicolumn{3}{c|}{68} & \multicolumn{3}{c|}{56} & \multicolumn{3}{c}{88} \\
\textbf{DoorOpen}                 & \multicolumn{3}{c|}{100} & \multicolumn{3}{c|}{100}  & \multicolumn{3}{c|}{100} & \multicolumn{3}{c|}{100} & \multicolumn{3}{c|}{100} & \multicolumn{3}{c|}{100} & \multicolumn{3}{c|}{100} & \multicolumn{3}{c|}{96} & \multicolumn{3}{c}{100} \\
\textbf{DoorUnlock}               & \multicolumn{3}{c|}{092} & \multicolumn{3}{c|}{100}  & \multicolumn{3}{c|}{100} & \multicolumn{3}{c|}{100} & \multicolumn{3}{c|}{100} & \multicolumn{3}{c|}{100} & \multicolumn{3}{c|}{100} & \multicolumn{3}{c|}{88} & \multicolumn{3}{c}{100} \\
\textbf{DrawerClose}              & \multicolumn{3}{c|}{100} & \multicolumn{3}{c|}{100}  & \multicolumn{3}{c|}{100} & \multicolumn{3}{c|}{100} & \multicolumn{3}{c|}{100} & \multicolumn{3}{c|}{100} & \multicolumn{3}{c|}{100} & \multicolumn{3}{c|}{100} & \multicolumn{3}{c}{100} \\
\textbf{DrawerOpen}               & \multicolumn{3}{c|}{100} & \multicolumn{3}{c|}{100}  & \multicolumn{3}{c|}{100} & \multicolumn{3}{c|}{100} & \multicolumn{3}{c|}{100} & \multicolumn{3}{c|}{100} & \multicolumn{3}{c|}{100} & \multicolumn{3}{c|}{100} & \multicolumn{3}{c}{100} \\
\textbf{FaucetClose}              & \multicolumn{3}{c|}{100} & \multicolumn{3}{c|}{96}   & \multicolumn{3}{c|}{96}  & \multicolumn{3}{c|}{36} & \multicolumn{3}{c|}{64} & \multicolumn{3}{c|}{100} & \multicolumn{3}{c|}{100} & \multicolumn{3}{c|}{100} & \multicolumn{3}{c}{100} \\
\textbf{FaucetOpen}               & \multicolumn{3}{c|}{100} & \multicolumn{3}{c|}{100}  & \multicolumn{3}{c|}{100} & \multicolumn{3}{c|}{76} & \multicolumn{3}{c|}{100} & \multicolumn{3}{c|}{100} & \multicolumn{3}{c|}{100} & \multicolumn{3}{c|}{100} & \multicolumn{3}{c}{100} \\
\textbf{HandlePress}              & \multicolumn{3}{c|}{100} & \multicolumn{3}{c|}{100}  & \multicolumn{3}{c|}{100} & \multicolumn{3}{c|}{100} & \multicolumn{3}{c|}{100} & \multicolumn{3}{c|}{100} & \multicolumn{3}{c|}{100} & \multicolumn{3}{c|}{100} & \multicolumn{3}{c}{100} \\
\textbf{HandlePull}               & \multicolumn{3}{c|}{60}  & \multicolumn{3}{c|}{48}   & \multicolumn{3}{c|}{48}  & \multicolumn{3}{c|}{52} & \multicolumn{3}{c|}{96} & \multicolumn{3}{c|}{64} & \multicolumn{3}{c|}{76} & \multicolumn{3}{c|}{56} & \multicolumn{3}{c}{96} \\
\textbf{HandlePressSide}          & \multicolumn{3}{c|}{92}  & \multicolumn{3}{c|}{68}   & \multicolumn{3}{c|}{68}  & \multicolumn{3}{c|}{100} & \multicolumn{3}{c|}{100} & \multicolumn{3}{c|}{100} & \multicolumn{3}{c|}{100} & \multicolumn{3}{c|}{100} & \multicolumn{3}{c}{100} \\
\textbf{HandlePullSide}           & \multicolumn{3}{c|}{12}  & \multicolumn{3}{c|}{0}    & \multicolumn{3}{c|}{0}   & \multicolumn{3}{c|}{0} & \multicolumn{3}{c|}{28} & \multicolumn{3}{c|}{16} & \multicolumn{3}{c|}{44} & \multicolumn{3}{c|}{24} & \multicolumn{3}{c}{100} \\
\textbf{LeverPull}                & \multicolumn{3}{c|}{52}  & \multicolumn{3}{c|}{16}   & \multicolumn{3}{c|}{16}  & \multicolumn{3}{c|}{32} & \multicolumn{3}{c|}{60} & \multicolumn{3}{c|}{44} & \multicolumn{3}{c|}{72} & \multicolumn{3}{c|}{92} & \multicolumn{3}{c}{88} \\
\textbf{PlateSlide}               & \multicolumn{3}{c|}{92}  & \multicolumn{3}{c|}{100}  & \multicolumn{3}{c|}{100} & \multicolumn{3}{c|}{100} & \multicolumn{3}{c|}{100} & \multicolumn{3}{c|}{100} & \multicolumn{3}{c|}{100} & \multicolumn{3}{c|}{100} & \multicolumn{3}{c}{100} \\
\textbf{PlateSlideBack}           & \multicolumn{3}{c|}{100} & \multicolumn{3}{c|}{100}  & \multicolumn{3}{c|}{100} & \multicolumn{3}{c|}{100} & \multicolumn{3}{c|}{84} & \multicolumn{3}{c|}{100} & \multicolumn{3}{c|}{100} & \multicolumn{3}{c|}{100} & \multicolumn{3}{c}{100} \\
\textbf{PlateSlideBackSide}       & \multicolumn{3}{c|}{100} & \multicolumn{3}{c|}{100}  & \multicolumn{3}{c|}{100} & \multicolumn{3}{c|}{100} & \multicolumn{3}{c|}{0} & \multicolumn{3}{c|}{100} & \multicolumn{3}{c|}{100} & \multicolumn{3}{c|}{100} & \multicolumn{3}{c}{100} \\
\textbf{PlateSlideSide}           & \multicolumn{3}{c|}{100} & \multicolumn{3}{c|}{100}  & \multicolumn{3}{c|}{100} & \multicolumn{3}{c|}{100} & \multicolumn{3}{c|}{100} & \multicolumn{3}{c|}{100} & \multicolumn{3}{c|}{100} & \multicolumn{3}{c|}{100} & \multicolumn{3}{c}{100} \\
\textbf{Reach}                    & \multicolumn{3}{c|}{48}  & \multicolumn{3}{c|}{40}   & \multicolumn{3}{c|}{56}  & \multicolumn{3}{c|}{36} & \multicolumn{3}{c|}{36} & \multicolumn{3}{c|}{28} & \multicolumn{3}{c|}{36} & \multicolumn{3}{c|}{28} & \multicolumn{3}{c}{32} \\
\textbf{ReachWall}                & \multicolumn{3}{c|}{76}  & \multicolumn{3}{c|}{64}   & \multicolumn{3}{c|}{64}  & \multicolumn{3}{c|}{68} & \multicolumn{3}{c|}{52} & \multicolumn{3}{c|}{76} & \multicolumn{3}{c|}{76} & \multicolumn{3}{c|}{72} & \multicolumn{3}{c}{80} \\
\textbf{WindowClose}              & \multicolumn{3}{c|}{100} & \multicolumn{3}{c|}{100}  & \multicolumn{3}{c|}{100} & \multicolumn{3}{c|}{16} & \multicolumn{3}{c|}{84} & \multicolumn{3}{c|}{100} & \multicolumn{3}{c|}{100} & \multicolumn{3}{c|}{100} & \multicolumn{3}{c}{100} \\
\textbf{WindowOpen}               & \multicolumn{3}{c|}{80}  & \multicolumn{3}{c|}{96}   & \multicolumn{3}{c|}{96}  & \multicolumn{3}{c|}{44} & \multicolumn{3}{c|}{92} & \multicolumn{3}{c|}{96} & \multicolumn{3}{c|}{100} & \multicolumn{3}{c|}{100} & \multicolumn{3}{c}{100} \\
\textbf{PegUnplugSide}            & \multicolumn{3}{c|}{56}  & \multicolumn{3}{c|}{48}   & \multicolumn{3}{c|}{48}  & \multicolumn{3}{c|}{80} & \multicolumn{3}{c|}{88} & \multicolumn{3}{c|}{76} & \multicolumn{3}{c|}{92} & \multicolumn{3}{c|}{72} & \multicolumn{3}{c}{100} \\
\textbf{Basketball}               & \multicolumn{3}{c|}{4}   & \multicolumn{3}{c|}{36}   & \multicolumn{3}{c|}{36}  & \multicolumn{3}{c|}{0} & \multicolumn{3}{c|}{0} & \multicolumn{3}{c|}{4} & \multicolumn{3}{c|}{8} & \multicolumn{3}{c|}{28} & \multicolumn{3}{c}{40} \\
\textbf{BinPicking}               & \multicolumn{3}{c|}{76}  & \multicolumn{3}{c|}{60}   & \multicolumn{3}{c|}{60}  & \multicolumn{3}{c|}{20} & \multicolumn{3}{c|}{72} & \multicolumn{3}{c|}{40} & \multicolumn{3}{c|}{52} & \multicolumn{3}{c|}{76} & \multicolumn{3}{c}{92} \\
\textbf{BoxClose}                 & \multicolumn{3}{c|}{48}  & \multicolumn{3}{c|}{72}   & \multicolumn{3}{c|}{72}  & \multicolumn{3}{c|}{24} & \multicolumn{3}{c|}{92} & \multicolumn{3}{c|}{68} & \multicolumn{3}{c|}{76} & \multicolumn{3}{c|}{72} & \multicolumn{3}{c}{64} \\
\textbf{CoffeePull}               & \multicolumn{3}{c|}{64}  & \multicolumn{3}{c|}{84}   & \multicolumn{3}{c|}{84}  & \multicolumn{3}{c|}{8} & \multicolumn{3}{c|}{80} & \multicolumn{3}{c|}{68} & \multicolumn{3}{c|}{76} & \multicolumn{3}{c|}{72} & \multicolumn{3}{c}{76} \\
\textbf{CoffeePush}               & \multicolumn{3}{c|}{64}  & \multicolumn{3}{c|}{88}   & \multicolumn{3}{c|}{84}  & \multicolumn{3}{c|}{100} & \multicolumn{3}{c|}{96} & \multicolumn{3}{c|}{76} & \multicolumn{3}{c|}{80} & \multicolumn{3}{c|}{68} & \multicolumn{3}{c}{92} \\
\textbf{Hammer}                   & \multicolumn{3}{c|}{68}  & \multicolumn{3}{c|}{8}    & \multicolumn{3}{c|}{8}   & \multicolumn{3}{c|}{92} & \multicolumn{3}{c|}{84} & \multicolumn{3}{c|}{100} & \multicolumn{3}{c|}{92} & \multicolumn{3}{c|}{88} & \multicolumn{3}{c}{100} \\
\textbf{PegInsertSide}            & \multicolumn{3}{c|}{32}  & \multicolumn{3}{c|}{28}   & \multicolumn{3}{c|}{28}  & \multicolumn{3}{c|}{16} & \multicolumn{3}{c|}{12} & \multicolumn{3}{c|}{32} & \multicolumn{3}{c|}{40} & \multicolumn{3}{c|}{44} & \multicolumn{3}{c}{60} \\
\textbf{PushWall}                 & \multicolumn{3}{c|}{56}  & \multicolumn{3}{c|}{52}   & \multicolumn{3}{c|}{52}  & \multicolumn{3}{c|}{24} & \multicolumn{3}{c|}{28} & \multicolumn{3}{c|}{64} & \multicolumn{3}{c|}{80} & \multicolumn{3}{c|}{24} & \multicolumn{3}{c}{72} \\
\textbf{Soccer}                   & \multicolumn{3}{c|}{8}   & \multicolumn{3}{c|}{24}   & \multicolumn{3}{c|}{24}  & \multicolumn{3}{c|}{20} & \multicolumn{3}{c|}{56} & \multicolumn{3}{c|}{12} & \multicolumn{3}{c|}{28} & \multicolumn{3}{c|}{28} & \multicolumn{3}{c}{36} \\
\textbf{Sweep}                    & \multicolumn{3}{c|}{32}  & \multicolumn{3}{c|}{40}   & \multicolumn{3}{c|}{40}  & \multicolumn{3}{c|}{36} & \multicolumn{3}{c|}{12} & \multicolumn{3}{c|}{60} & \multicolumn{3}{c|}{72} & \multicolumn{3}{c|}{60} & \multicolumn{3}{c}{76} \\
\textbf{SweepInto}                & \multicolumn{3}{c|}{72}  & \multicolumn{3}{c|}{52}   & \multicolumn{3}{c|}{52}  & \multicolumn{3}{c|}{36} & \multicolumn{3}{c|}{84} & \multicolumn{3}{c|}{68} & \multicolumn{3}{c|}{80} & \multicolumn{3}{c|}{68} & \multicolumn{3}{c}{88} \\
\textbf{Assembly}                 & \multicolumn{3}{c|}{56}  & \multicolumn{3}{c|}{88}   & \multicolumn{3}{c|}{88}  & \multicolumn{3}{c|}{100} & \multicolumn{3}{c|}{100} & \multicolumn{3}{c|}{80} & \multicolumn{3}{c|}{84} & \multicolumn{3}{c|}{84} & \multicolumn{3}{c}{96} \\
\textbf{HandInsert}               & \multicolumn{3}{c|}{44}  & \multicolumn{3}{c|}{60}   & \multicolumn{3}{c|}{60}  & \multicolumn{3}{c|}{4} & \multicolumn{3}{c|}{76} & \multicolumn{3}{c|}{48} & \multicolumn{3}{c|}{76} & \multicolumn{3}{c|}{76} & \multicolumn{3}{c}{76} \\
\textbf{PickOutOfHole}            & \multicolumn{3}{c|}{36}  & \multicolumn{3}{c|}{52}   & \multicolumn{3}{c|}{52}  & \multicolumn{3}{c|}{12} & \multicolumn{3}{c|}{0} & \multicolumn{3}{c|}{8} & \multicolumn{3}{c|}{12} & \multicolumn{3}{c|}{68} & \multicolumn{3}{c}{76} \\
\textbf{PickPlace}                & \multicolumn{3}{c|}{28}  & \multicolumn{3}{c|}{24}   & \multicolumn{3}{c|}{24}  & \multicolumn{3}{c|}{44} & \multicolumn{3}{c|}{84} & \multicolumn{3}{c|}{52} & \multicolumn{3}{c|}{56} & \multicolumn{3}{c|}{85} & \multicolumn{3}{c}{84} \\
\textbf{Push}                     & \multicolumn{3}{c|}{48}  & \multicolumn{3}{c|}{60}   & \multicolumn{3}{c|}{60}  & \multicolumn{3}{c|}{52} & \multicolumn{3}{c|}{36} & \multicolumn{3}{c|}{48} & \multicolumn{3}{c|}{68} & \multicolumn{3}{c|}{88} & \multicolumn{3}{c}{80} \\
\textbf{PushBack}                 & \multicolumn{3}{c|}{56}  & \multicolumn{3}{c|}{48}   & \multicolumn{3}{c|}{48}  & \multicolumn{3}{c|}{52} & \multicolumn{3}{c|}{64} & \multicolumn{3}{c|}{52} & \multicolumn{3}{c|}{56} & \multicolumn{3}{c|}{52} & \multicolumn{3}{c}{80} \\
\textbf{PickPlaceWall}            & \multicolumn{3}{c|}{28}  & \multicolumn{3}{c|}{24}   & \multicolumn{3}{c|}{24}  & \multicolumn{3}{c|}{20} & \multicolumn{3}{c|}{4} & \multicolumn{3}{c|}{20} & \multicolumn{3}{c|}{32} & \multicolumn{3}{c|}{12} & \multicolumn{3}{c}{40} \\
\textbf{StickPull}                & \multicolumn{3}{c|}{48}  & \multicolumn{3}{c|}{80}   & \multicolumn{3}{c|}{80}  & \multicolumn{3}{c|}{24} & \multicolumn{3}{c|}{60} & \multicolumn{3}{c|}{64} & \multicolumn{3}{c|}{80} & \multicolumn{3}{c|}{80} & \multicolumn{3}{c}{96} \\
\textbf{StickPush}                & \multicolumn{3}{c|}{84}  & \multicolumn{3}{c|}{72}   & \multicolumn{3}{c|}{72}  & \multicolumn{3}{c|}{44} & \multicolumn{3}{c|}{20} & \multicolumn{3}{c|}{80} & \multicolumn{3}{c|}{100} & \multicolumn{3}{c|}{72} & \multicolumn{3}{c}{100} \\
\textbf{ShelfPlace}               & \multicolumn{3}{c|}{12}  & \multicolumn{3}{c|}{12}   & \multicolumn{3}{c|}{12}  & \multicolumn{3}{c|}{32} & \multicolumn{3}{c|}{40} & \multicolumn{3}{c|}{8} & \multicolumn{3}{c|}{8} & \multicolumn{3}{c|}{8} & \multicolumn{3}{c}{36} \\
\textbf{Disassemble}              & \multicolumn{3}{c|}{60}  & \multicolumn{3}{c|}{4}    & \multicolumn{3}{c|}{4}   & \multicolumn{3}{c|}{28} & \multicolumn{3}{c|}{16} & \multicolumn{3}{c|}{60} & \multicolumn{3}{c|}{72} & \multicolumn{3}{c|}{60} & \multicolumn{3}{c}{76} \\

\bottomrule
\end{tabular}
\end{adjustbox}
\end{table*}

\begin{table*}[htbp]
\centering
\caption{All multi-task results on RLBench.}
\begin{adjustbox}{width=\linewidth, totalheight=\textheight, keepaspectratio, scale=0.98}
\scriptsize 

\newcommand{\methodcolwidth}{1.7em}   
\newcommand{\taskcolwidth}{7.0em}    

\begin{tabular}{
    >{\raggedright\arraybackslash\hspace{0pt}\scriptsize}p{\taskcolwidth}  
    *{9}{ 
        |>{\centering\arraybackslash}p{\methodcolwidth}
         >{\centering\arraybackslash}p{\methodcolwidth}
         >{\centering\arraybackslash}p{\methodcolwidth}
    }
}
\toprule
\multirow{2}{*}{\textbf{Method}} &
\multicolumn{3}{c|}{\multirow{2}{*}{\textbf{CLIP}}} &
\multicolumn{3}{c|}{\multirow{2}{*}{\textbf{R3M}}} &
\multicolumn{3}{c|}{\multirow{2}{*}{\textbf{VC-1}}} &
\multicolumn{3}{c|}{\multirow{2}{*}{\textbf{DP3}}} &
\multicolumn{3}{c|}{\multirow{2}{*}{\textbf{Lift3d}}} &
\multicolumn{3}{c|}{\multirow{2}{*}{\textbf{SPA-S}}} &
\multicolumn{3}{c|}{\multirow{2}{*}{\textbf{SPA}}} &
\multicolumn{3}{c|}{\multirow{2}{*}{\textbf{GP3-S}}} &
\multicolumn{3}{c}{\multirow{2}{*}{\textbf{GP3}}} \\

& & & & & & & & & & & & & & & & & & & & & & & & & & & \\
\midrule
\textbf{CloseBox}       & \multicolumn{3}{c|}{68} & \multicolumn{3}{c|}{92} & \multicolumn{3}{c|}{80} & \multicolumn{3}{c|}{96} & \multicolumn{3}{c|}{92} & \multicolumn{3}{c|}{80} & \multicolumn{3}{c|}{92} &\multicolumn{3}{c|}{100} & \multicolumn{3}{c}{100} \\
\textbf{PutRubbishInBin}   & \multicolumn{3}{c|}{8} & \multicolumn{3}{c|}{12} & \multicolumn{3}{c|}{24} & \multicolumn{3}{c|}{12} & \multicolumn{3}{c|}{40} & \multicolumn{3}{c|}{4} & \multicolumn{3}{c|}{0} &\multicolumn{3}{c|}{88} & \multicolumn{3}{c}{80} \\
\textbf{CloseLaptopLid}          & \multicolumn{3}{c|}{84} & \multicolumn{3}{c|}{96} & \multicolumn{3}{c|}{84} & \multicolumn{3}{c|}{72} & \multicolumn{3}{c|}{84} & \multicolumn{3}{c|}{92} & \multicolumn{3}{c|}{72} &\multicolumn{3}{c|}{96} & \multicolumn{3}{c}{96} \\
\textbf{WaterPlants}             & \multicolumn{3}{c|}{32} & \multicolumn{3}{c|}{36} & \multicolumn{3}{c|}{32} & \multicolumn{3}{c|}{4} & \multicolumn{3}{c|}{8} & \multicolumn{3}{c|}{32} & \multicolumn{3}{c|}{24} &\multicolumn{3}{c|}{36} & \multicolumn{3}{c}{52} \\
\textbf{UnplugCharger}                & \multicolumn{3}{c|}{28} & \multicolumn{3}{c|}{20} & \multicolumn{3}{c|}{28} & \multicolumn{3}{c|}{40} & \multicolumn{3}{c|}{42} & \multicolumn{3}{c|}{28} & \multicolumn{3}{c|}{20} &\multicolumn{3}{c|}{60} & \multicolumn{3}{c}{72} \\
\textbf{ToiletSeatDown}                 & \multicolumn{3}{c|}{96} & \multicolumn{3}{c|}{100} & \multicolumn{3}{c|}{100} & \multicolumn{3}{c|}{80} & \multicolumn{3}{c|}{88} & \multicolumn{3}{c|}{100} & \multicolumn{3}{c|}{92} &\multicolumn{3}{c|}{100} & \multicolumn{3}{c}{100} \\
\bottomrule
\end{tabular}
\end{adjustbox}
\end{table*}

\begin{table*}[htbp]
\centering
\caption{All multi-task results on Meta-World.}
\begin{adjustbox}{width=\linewidth, totalheight=\textheight, keepaspectratio, scale=0.98}
\scriptsize 

\newcommand{\methodcolwidth}{1.7em}   
\newcommand{\taskcolwidth}{10.0em}    

\begin{tabular}{
    >{\raggedright\arraybackslash\hspace{0pt}\scriptsize}p{\taskcolwidth}  
    *{9}{ 
        |>{\centering\arraybackslash}p{\methodcolwidth}
         >{\centering\arraybackslash}p{\methodcolwidth}
         >{\centering\arraybackslash}p{\methodcolwidth}
    }
}
\toprule
\multirow{2}{*}{\textbf{Method}} &
\multicolumn{3}{c|}{\multirow{2}{*}{\textbf{CLIP}}} &
\multicolumn{3}{c|}{\multirow{2}{*}{\textbf{R3M}}} &
\multicolumn{3}{c|}{\multirow{2}{*}{\textbf{VC-1}}} &
\multicolumn{3}{c|}{\multirow{2}{*}{\textbf{DP3}}} &
\multicolumn{3}{c|}{\multirow{2}{*}{\textbf{Lift3d}}} &
\multicolumn{3}{c|}{\multirow{2}{*}{\textbf{SPA-S}}} &
\multicolumn{3}{c|}{\multirow{2}{*}{\textbf{SPA}}} &
\multicolumn{3}{c|}{\multirow{2}{*}{\textbf{GP3-S}}} &
\multicolumn{3}{c}{\multirow{2}{*}{\textbf{GP3}}} \\

& & & & & & & & & & & & & & & & & & & & & & & & & & & \\
\midrule
\textbf{ButtonPress}              & \multicolumn{3}{c|}{100} & \multicolumn{3}{c|}{100} & \multicolumn{3}{c|}{100} & \multicolumn{3}{c|}{100} & \multicolumn{3}{c|}{100} & \multicolumn{3}{c|}{100} & \multicolumn{3}{c|}{100} & \multicolumn{3}{c|}{100} & \multicolumn{3}{c}{100} \\
\textbf{ButtonPressTopdown}       & \multicolumn{3}{c|}{72}  & \multicolumn{3}{c|}{88}  & \multicolumn{3}{c|}{100} & \multicolumn{3}{c|}{92}  & \multicolumn{3}{c|}{100} & \multicolumn{3}{c|}{88}  & \multicolumn{3}{c|}{100} & \multicolumn{3}{c|}{100} & \multicolumn{3}{c}{100} \\
\textbf{ButtonPressTopdownWall}   & \multicolumn{3}{c|}{96}  & \multicolumn{3}{c|}{96}  & \multicolumn{3}{c|}{100} & \multicolumn{3}{c|}{8}   & \multicolumn{3}{c|}{100} & \multicolumn{3}{c|}{100} & \multicolumn{3}{c|}{100} & \multicolumn{3}{c|}{100} & \multicolumn{3}{c}{100} \\
\textbf{ButtonPressWall}          & \multicolumn{3}{c|}{96}  & \multicolumn{3}{c|}{100} & \multicolumn{3}{c|}{96}  & \multicolumn{3}{c|}{84}  & \multicolumn{3}{c|}{100} & \multicolumn{3}{c|}{100} & \multicolumn{3}{c|}{100} & \multicolumn{3}{c|}{96}  & \multicolumn{3}{c}{100} \\
\textbf{CoffeeButton}             & \multicolumn{3}{c|}{100} & \multicolumn{3}{c|}{100} & \multicolumn{3}{c|}{100} & \multicolumn{3}{c|}{100} & \multicolumn{3}{c|}{100} & \multicolumn{3}{c|}{100} & \multicolumn{3}{c|}{100} & \multicolumn{3}{c|}{100} & \multicolumn{3}{c}{100} \\
\textbf{DialTurn}                 & \multicolumn{3}{c|}{52}  & \multicolumn{3}{c|}{24}  & \multicolumn{3}{c|}{32}  & \multicolumn{3}{c|}{28}  & \multicolumn{3}{c|}{12}  & \multicolumn{3}{c|}{32}  & \multicolumn{3}{c|}{64}  & \multicolumn{3}{c|}{84}  & \multicolumn{3}{c}{76} \\
\textbf{DoorClose}                & \multicolumn{3}{c|}{100} & \multicolumn{3}{c|}{100} & \multicolumn{3}{c|}{100} & \multicolumn{3}{c|}{100} & \multicolumn{3}{c|}{100} & \multicolumn{3}{c|}{100} & \multicolumn{3}{c|}{100} & \multicolumn{3}{c|}{100} & \multicolumn{3}{c}{100} \\
\textbf{DoorLock}                 & \multicolumn{3}{c|}{48}  & \multicolumn{3}{c|}{64}  & \multicolumn{3}{c|}{44}  & \multicolumn{3}{c|}{4}   & \multicolumn{3}{c|}{88}  & \multicolumn{3}{c|}{60}  & \multicolumn{3}{c|}{64}  & \multicolumn{3}{c|}{72}  & \multicolumn{3}{c}{76} \\
\textbf{DoorOpen}                 & \multicolumn{3}{c|}{100} & \multicolumn{3}{c|}{100} & \multicolumn{3}{c|}{100} & \multicolumn{3}{c|}{96}  & \multicolumn{3}{c|}{100} & \multicolumn{3}{c|}{100} & \multicolumn{3}{c|}{100} & \multicolumn{3}{c|}{96}  & \multicolumn{3}{c}{100} \\
\textbf{DoorUnlock}               & \multicolumn{3}{c|}{96}  & \multicolumn{3}{c|}{100} & \multicolumn{3}{c|}{96}  & \multicolumn{3}{c|}{56}  & \multicolumn{3}{c|}{96}  & \multicolumn{3}{c|}{100} & \multicolumn{3}{c|}{100} & \multicolumn{3}{c|}{96}  & \multicolumn{3}{c}{92} \\
\textbf{DrawerClose}              & \multicolumn{3}{c|}{96}  & \multicolumn{3}{c|}{100} & \multicolumn{3}{c|}{100} & \multicolumn{3}{c|}{96}  & \multicolumn{3}{c|}{100} & \multicolumn{3}{c|}{100} & \multicolumn{3}{c|}{100} & \multicolumn{3}{c|}{100} & \multicolumn{3}{c}{100} \\
\textbf{DrawerOpen}               & \multicolumn{3}{c|}{96}  & \multicolumn{3}{c|}{100} & \multicolumn{3}{c|}{100} & \multicolumn{3}{c|}{80}  & \multicolumn{3}{c|}{100} & \multicolumn{3}{c|}{100} & \multicolumn{3}{c|}{100} & \multicolumn{3}{c|}{100} & \multicolumn{3}{c}{100} \\
\textbf{FaucetClose}              & \multicolumn{3}{c|}{76}  & \multicolumn{3}{c|}{76}  & \multicolumn{3}{c|}{80}  & \multicolumn{3}{c|}{48}  & \multicolumn{3}{c|}{36}  & \multicolumn{3}{c|}{64}  & \multicolumn{3}{c|}{92}  & \multicolumn{3}{c|}{100} & \multicolumn{3}{c}{100} \\
\textbf{FaucetOpen}               & \multicolumn{3}{c|}{100} & \multicolumn{3}{c|}{100} & \multicolumn{3}{c|}{100} & \multicolumn{3}{c|}{64}  & \multicolumn{3}{c|}{0}   & \multicolumn{3}{c|}{100} & \multicolumn{3}{c|}{100} & \multicolumn{3}{c|}{100} & \multicolumn{3}{c}{100} \\
\textbf{HandlePress}              & \multicolumn{3}{c|}{100} & \multicolumn{3}{c|}{100} & \multicolumn{3}{c|}{100} & \multicolumn{3}{c|}{100} & \multicolumn{3}{c|}{100} & \multicolumn{3}{c|}{92}  & \multicolumn{3}{c|}{100} & \multicolumn{3}{c|}{100} & \multicolumn{3}{c}{100} \\
\textbf{HandlePull}               & \multicolumn{3}{c|}{44}  & \multicolumn{3}{c|}{76}  & \multicolumn{3}{c|}{44}  & \multicolumn{3}{c|}{12}  & \multicolumn{3}{c|}{100} & \multicolumn{3}{c|}{48}  & \multicolumn{3}{c|}{72}  & \multicolumn{3}{c|}{76}  & \multicolumn{3}{c}{72} \\
\textbf{HandlePressSide}          & \multicolumn{3}{c|}{92}  & \multicolumn{3}{c|}{100} & \multicolumn{3}{c|}{80}  & \multicolumn{3}{c|}{48}  & \multicolumn{3}{c|}{100} & \multicolumn{3}{c|}{96}  & \multicolumn{3}{c|}{100} & \multicolumn{3}{c|}{100} & \multicolumn{3}{c}{100} \\
\textbf{HandlePullSide}           & \multicolumn{3}{c|}{20}  & \multicolumn{3}{c|}{48}  & \multicolumn{3}{c|}{16}  & \multicolumn{3}{c|}{0}   & \multicolumn{3}{c|}{68}  & \multicolumn{3}{c|}{20}  & \multicolumn{3}{c|}{36}  & \multicolumn{3}{c|}{100} & \multicolumn{3}{c}{40} \\
\textbf{LeverPull}                & \multicolumn{3}{c|}{52}  & \multicolumn{3}{c|}{56}  & \multicolumn{3}{c|}{64}  & \multicolumn{3}{c|}{64}  & \multicolumn{3}{c|}{88}  & \multicolumn{3}{c|}{56}  & \multicolumn{3}{c|}{52}  & \multicolumn{3}{c|}{76}  & \multicolumn{3}{c}{52} \\
\textbf{PlateSlide}               & \multicolumn{3}{c|}{92}  & \multicolumn{3}{c|}{100} & \multicolumn{3}{c|}{100} & \multicolumn{3}{c|}{28}  & \multicolumn{3}{c|}{0}   & \multicolumn{3}{c|}{96}  & \multicolumn{3}{c|}{100} & \multicolumn{3}{c|}{100} & \multicolumn{3}{c}{100} \\
\textbf{PlateSlideBack}           & \multicolumn{3}{c|}{96}  & \multicolumn{3}{c|}{44}  & \multicolumn{3}{c|}{100} & \multicolumn{3}{c|}{0}   & \multicolumn{3}{c|}{0}   & \multicolumn{3}{c|}{68}  & \multicolumn{3}{c|}{100} & \multicolumn{3}{c|}{0}   & \multicolumn{3}{c}{0} \\
\textbf{PlateSlideBackSide}       & \multicolumn{3}{c|}{100} & \multicolumn{3}{c|}{100} & \multicolumn{3}{c|}{100} & \multicolumn{3}{c|}{4}   & \multicolumn{3}{c|}{8}   & \multicolumn{3}{c|}{100} & \multicolumn{3}{c|}{100} & \multicolumn{3}{c|}{100} & \multicolumn{3}{c}{100} \\
\textbf{PlateSlideSide}           & \multicolumn{3}{c|}{100} & \multicolumn{3}{c|}{100} & \multicolumn{3}{c|}{100} & \multicolumn{3}{c|}{92}  & \multicolumn{3}{c|}{100} & \multicolumn{3}{c|}{100} & \multicolumn{3}{c|}{100} & \multicolumn{3}{c|}{100} & \multicolumn{3}{c}{100} \\
\textbf{Reach}                    & \multicolumn{3}{c|}{32}  & \multicolumn{3}{c|}{24}  & \multicolumn{3}{c|}{32}  & \multicolumn{3}{c|}{16}  & \multicolumn{3}{c|}{32}  & \multicolumn{3}{c|}{32}  & \multicolumn{3}{c|}{40}  & \multicolumn{3}{c|}{32}  & \multicolumn{3}{c}{56} \\
\textbf{ReachWall}                & \multicolumn{3}{c|}{72}  & \multicolumn{3}{c|}{80}  & \multicolumn{3}{c|}{68}  & \multicolumn{3}{c|}{60}  & \multicolumn{3}{c|}{36}  & \multicolumn{3}{c|}{80}  & \multicolumn{3}{c|}{60}  & \multicolumn{3}{c|}{72}  & \multicolumn{3}{c}{88} \\
\textbf{WindowClose}              & \multicolumn{3}{c|}{100} & \multicolumn{3}{c|}{100} & \multicolumn{3}{c|}{100} & \multicolumn{3}{c|}{88}  & \multicolumn{3}{c|}{100} & \multicolumn{3}{c|}{100} & \multicolumn{3}{c|}{100} & \multicolumn{3}{c|}{100} & \multicolumn{3}{c}{100} \\
\textbf{WindowOpen}               & \multicolumn{3}{c|}{92}  & \multicolumn{3}{c|}{96}  & \multicolumn{3}{c|}{96}  & \multicolumn{3}{c|}{4}   & \multicolumn{3}{c|}{80}  & \multicolumn{3}{c|}{100} & \multicolumn{3}{c|}{96}  & \multicolumn{3}{c|}{92}  & \multicolumn{3}{c}{100} \\
\textbf{PegUnplugSide}            & \multicolumn{3}{c|}{36}  & \multicolumn{3}{c|}{76}  & \multicolumn{3}{c|}{52}  & \multicolumn{3}{c|}{8}   & \multicolumn{3}{c|}{68}  & \multicolumn{3}{c|}{56}  & \multicolumn{3}{c|}{56}  & \multicolumn{3}{c|}{68}  & \multicolumn{3}{c}{52} \\
\textbf{Basketball}               & \multicolumn{3}{c|}{12}  & \multicolumn{3}{c|}{60}  & \multicolumn{3}{c|}{12}  & \multicolumn{3}{c|}{0}   & \multicolumn{3}{c|}{0}   & \multicolumn{3}{c|}{20}  & \multicolumn{3}{c|}{16}  & \multicolumn{3}{c|}{68}  & \multicolumn{3}{c}{72} \\
\textbf{BinPicking}               & \multicolumn{3}{c|}{80}  & \multicolumn{3}{c|}{0}   & \multicolumn{3}{c|}{72}  & \multicolumn{3}{c|}{8}   & \multicolumn{3}{c|}{0}   & \multicolumn{3}{c|}{52}  & \multicolumn{3}{c|}{52}  & \multicolumn{3}{c|}{88}  & \multicolumn{3}{c}{76} \\
\textbf{BoxClose}                 & \multicolumn{3}{c|}{64}  & \multicolumn{3}{c|}{56}  & \multicolumn{3}{c|}{76}  & \multicolumn{3}{c|}{0}   & \multicolumn{3}{c|}{0}   & \multicolumn{3}{c|}{64}  & \multicolumn{3}{c|}{76}  & \multicolumn{3}{c|}{64}  & \multicolumn{3}{c}{68} \\
\textbf{CoffeePull}               & \multicolumn{3}{c|}{64}  & \multicolumn{3}{c|}{88}  & \multicolumn{3}{c|}{76}  & \multicolumn{3}{c|}{12}  & \multicolumn{3}{c|}{84}  & \multicolumn{3}{c|}{84}  & \multicolumn{3}{c|}{64}  & \multicolumn{3}{c|}{96}  & \multicolumn{3}{c}{96} \\
\textbf{CoffeePush}               & \multicolumn{3}{c|}{48}  & \multicolumn{3}{c|}{64}  & \multicolumn{3}{c|}{24}  & \multicolumn{3}{c|}{24}  & \multicolumn{3}{c|}{96}  & \multicolumn{3}{c|}{48}  & \multicolumn{3}{c|}{56}  & \multicolumn{3}{c|}{96}  & \multicolumn{3}{c}{100} \\
\textbf{Hammer}                   & \multicolumn{3}{c|}{60}  & \multicolumn{3}{c|}{100} & \multicolumn{3}{c|}{80}  & \multicolumn{3}{c|}{24}  & \multicolumn{3}{c|}{36}  & \multicolumn{3}{c|}{84}  & \multicolumn{3}{c|}{84}  & \multicolumn{3}{c|}{88}  & \multicolumn{3}{c}{72} \\
\textbf{PegInsertSide}            & \multicolumn{3}{c|}{24}  & \multicolumn{3}{c|}{48}  & \multicolumn{3}{c|}{28}  & \multicolumn{3}{c|}{40}  & \multicolumn{3}{c|}{32}  & \multicolumn{3}{c|}{32}  & \multicolumn{3}{c|}{52}  & \multicolumn{3}{c|}{56}  & \multicolumn{3}{c}{68} \\
\textbf{PushWall}                 & \multicolumn{3}{c|}{44}  & \multicolumn{3}{c|}{52}  & \multicolumn{3}{c|}{60}  & \multicolumn{3}{c|}{0}   & \multicolumn{3}{c|}{28}  & \multicolumn{3}{c|}{36}  & \multicolumn{3}{c|}{92}  & \multicolumn{3}{c|}{92}  & \multicolumn{3}{c}{88} \\
\textbf{Soccer}                   & \multicolumn{3}{c|}{20}  & \multicolumn{3}{c|}{20}  & \multicolumn{3}{c|}{8}   & \multicolumn{3}{c|}{28}  & \multicolumn{3}{c|}{44}  & \multicolumn{3}{c|}{12}  & \multicolumn{3}{c|}{28}  & \multicolumn{3}{c|}{36}  & \multicolumn{3}{c}{32} \\
\textbf{Sweep}                  & \multicolumn{3}{c|}{48}  & \multicolumn{3}{c|}{60}  & \multicolumn{3}{c|}{56}  & \multicolumn{3}{c|}{16}  & \multicolumn{3}{c|}{12}  & \multicolumn{3}{c|}{72}  & \multicolumn{3}{c|}{80}  & \multicolumn{3}{c|}{92}  & \multicolumn{3}{c}{96} \\
\textbf{SweepInto}                & \multicolumn{3}{c|}{72}  & \multicolumn{3}{c|}{28}  & \multicolumn{3}{c|}{68}  & \multicolumn{3}{c|}{0}   & \multicolumn{3}{c|}{12}  & \multicolumn{3}{c|}{72}  & \multicolumn{3}{c|}{68}  & \multicolumn{3}{c|}{88}  & \multicolumn{3}{c}{84} \\
\textbf{Assembly}                 & \multicolumn{3}{c|}{32}  & \multicolumn{3}{c|}{72}  & \multicolumn{3}{c|}{52}  & \multicolumn{3}{c|}{0}   & \multicolumn{3}{c|}{0}   & \multicolumn{3}{c|}{40}  & \multicolumn{3}{c|}{68}  & \multicolumn{3}{c|}{92}  & \multicolumn{3}{c}{88} \\
\textbf{HandInsert}               & \multicolumn{3}{c|}{40}  & \multicolumn{3}{c|}{32}  & \multicolumn{3}{c|}{56}  & \multicolumn{3}{c|}{0}   & \multicolumn{3}{c|}{0}   & \multicolumn{3}{c|}{48}  & \multicolumn{3}{c|}{44}  & \multicolumn{3}{c|}{76}  & \multicolumn{3}{c}{84} \\
\textbf{PickOutOfHole}            & \multicolumn{3}{c|}{64}  & \multicolumn{3}{c|}{40}  & \multicolumn{3}{c|}{24}  & \multicolumn{3}{c|}{0}   & \multicolumn{3}{c|}{0}   & \multicolumn{3}{c|}{44}  & \multicolumn{3}{c|}{44}  & \multicolumn{3}{c|}{92}  & \multicolumn{3}{c}{100} \\
\textbf{PickPlace}                & \multicolumn{3}{c|}{24}  & \multicolumn{3}{c|}{24}  & \multicolumn{3}{c|}{44}  & \multicolumn{3}{c|}{20}  & \multicolumn{3}{c|}{60}  & \multicolumn{3}{c|}{36}  & \multicolumn{3}{c|}{48}  & \multicolumn{3}{c|}{68}  & \multicolumn{3}{c}{92} \\
\textbf{Push}                     & \multicolumn{3}{c|}{24}  & \multicolumn{3}{c|}{44}  & \multicolumn{3}{c|}{28}  & \multicolumn{3}{c|}{16}  & \multicolumn{3}{c|}{64}  & \multicolumn{3}{c|}{44}  & \multicolumn{3}{c|}{56}  & \multicolumn{3}{c|}{76}  & \multicolumn{3}{c}{100} \\
\textbf{PushBack}                 & \multicolumn{3}{c|}{48}  & \multicolumn{3}{c|}{28}  & \multicolumn{3}{c|}{28}  & \multicolumn{3}{c|}{24}  & \multicolumn{3}{c|}{32}  & \multicolumn{3}{c|}{36}  & \multicolumn{3}{c|}{52}  & \multicolumn{3}{c|}{52}   & \multicolumn{3}{c}{76} \\
\textbf{PickPlaceWall}            & \multicolumn{3}{c|}{32}  & \multicolumn{3}{c|}{32}  & \multicolumn{3}{c|}{32}  & \multicolumn{3}{c|}{4}   & \multicolumn{3}{c|}{24}  & \multicolumn{3}{c|}{16}  & \multicolumn{3}{c|}{48}  & \multicolumn{3}{c|}{72}  & \multicolumn{3}{c}{92} \\
\textbf{StickPull}                & \multicolumn{3}{c|}{48}  & \multicolumn{3}{c|}{72}  & \multicolumn{3}{c|}{48}  & \multicolumn{3}{c|}{4}   & \multicolumn{3}{c|}{44}  & \multicolumn{3}{c|}{64}  & \multicolumn{3}{c|}{48}  & \multicolumn{3}{c|}{60}  & \multicolumn{3}{c}{76} \\
\textbf{StickPush}                & \multicolumn{3}{c|}{88}  & \multicolumn{3}{c|}{100} & \multicolumn{3}{c|}{64}  & \multicolumn{3}{c|}{24}  & \multicolumn{3}{c|}{24}  & \multicolumn{3}{c|}{88}  & \multicolumn{3}{c|}{100} & \multicolumn{3}{c|}{88}  & \multicolumn{3}{c}{100} \\
\textbf{ShelfPlace}               & \multicolumn{3}{c|}{8}   & \multicolumn{3}{c|}{16}  & \multicolumn{3}{c|}{4}   & \multicolumn{3}{c|}{8}   & \multicolumn{3}{c|}{20}  & \multicolumn{3}{c|}{12}  & \multicolumn{3}{c|}{16}  & \multicolumn{3}{c|}{36}  & \multicolumn{3}{c}{52} \\
\textbf{Disassemble}              & \multicolumn{3}{c|}{12}  & \multicolumn{3}{c|}{0}   & \multicolumn{3}{c|}{12}  & \multicolumn{3}{c|}{16}  & \multicolumn{3}{c|}{24}  & \multicolumn{3}{c|}{36}  & \multicolumn{3}{c|}{48}  & \multicolumn{3}{c|}{20}  & \multicolumn{3}{c}{40} \\

\bottomrule
\end{tabular}
\end{adjustbox}
\end{table*}

\section{More Experiments on Real-World}
\label{sec:real-world-decieved}

\begin{figure}[htbp]
    \centering
    \captionsetup[subfigure]{justification=centering, singlelinecheck=false, font=small}  
    
    \begin{subfigure}[t]{\linewidth}
        \centering
        \includegraphics[width=\linewidth]{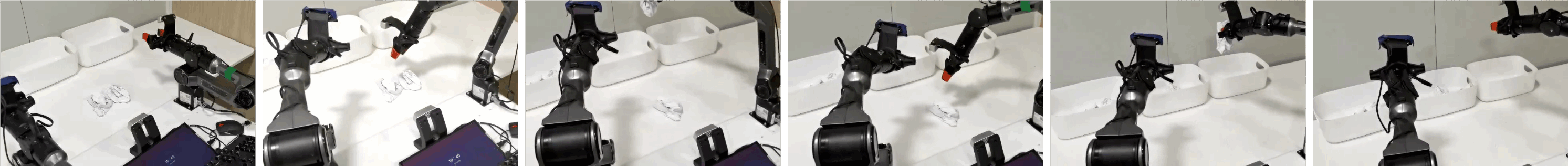}
        \caption{Catch and Throw: catch desktop paper ball and throw it away}
        \label{fig:catch_throw}
    \end{subfigure}
    
    \vspace{3mm}  
    
    \begin{subfigure}[t]{\linewidth}
        \centering
        \includegraphics[width=\linewidth]{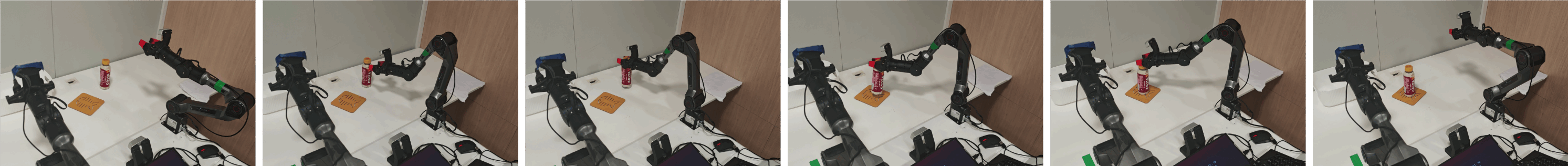}
        \caption{Grab Coffee: retrieve a coffee from the desk and put it on the coaster}
        \label{fig:grab_coffee}
    \end{subfigure}
    
    \vspace{3mm}
    
    \begin{subfigure}[t]{\linewidth}
        \centering
        \includegraphics[width=\linewidth]{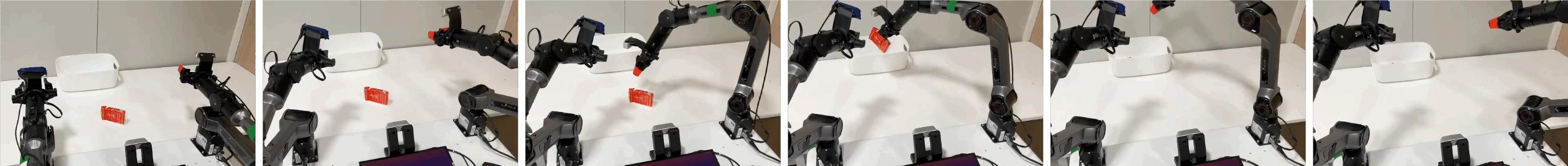}
        \caption{Clear Desk: clear items on desk}
        \label{fig:clear_desk}
    \end{subfigure}
    
    \vspace{3mm}
    
    \begin{subfigure}[t]{\linewidth}
        \centering
        \includegraphics[width=\linewidth]{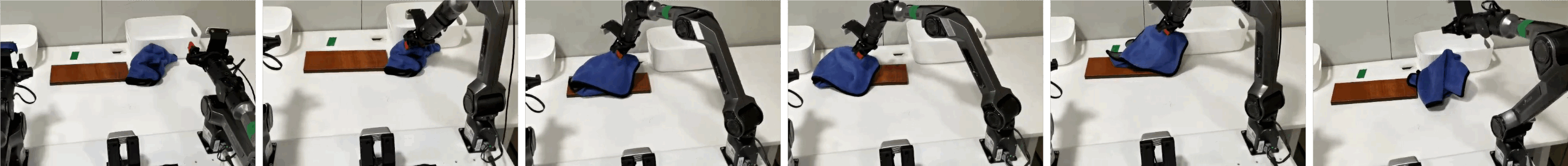}
        \caption{Clean Desktop: grasp the cloth and clean table}
        \label{fig:grab_cloth}
    \end{subfigure}

    \caption{Examples of base task executions with GP3.}
    \label{fig:gp3_base_tasks}
\end{figure}

In this section, we present additional results from real-world experiments to further validate GP3's 3D understanding. Figure~\ref{fig:gp3_base_tasks} illustrates the experimental setup used in Section~\ref{section:E_2}.

\begin{figure}[H]
    \centering
    \includegraphics[width=\linewidth]{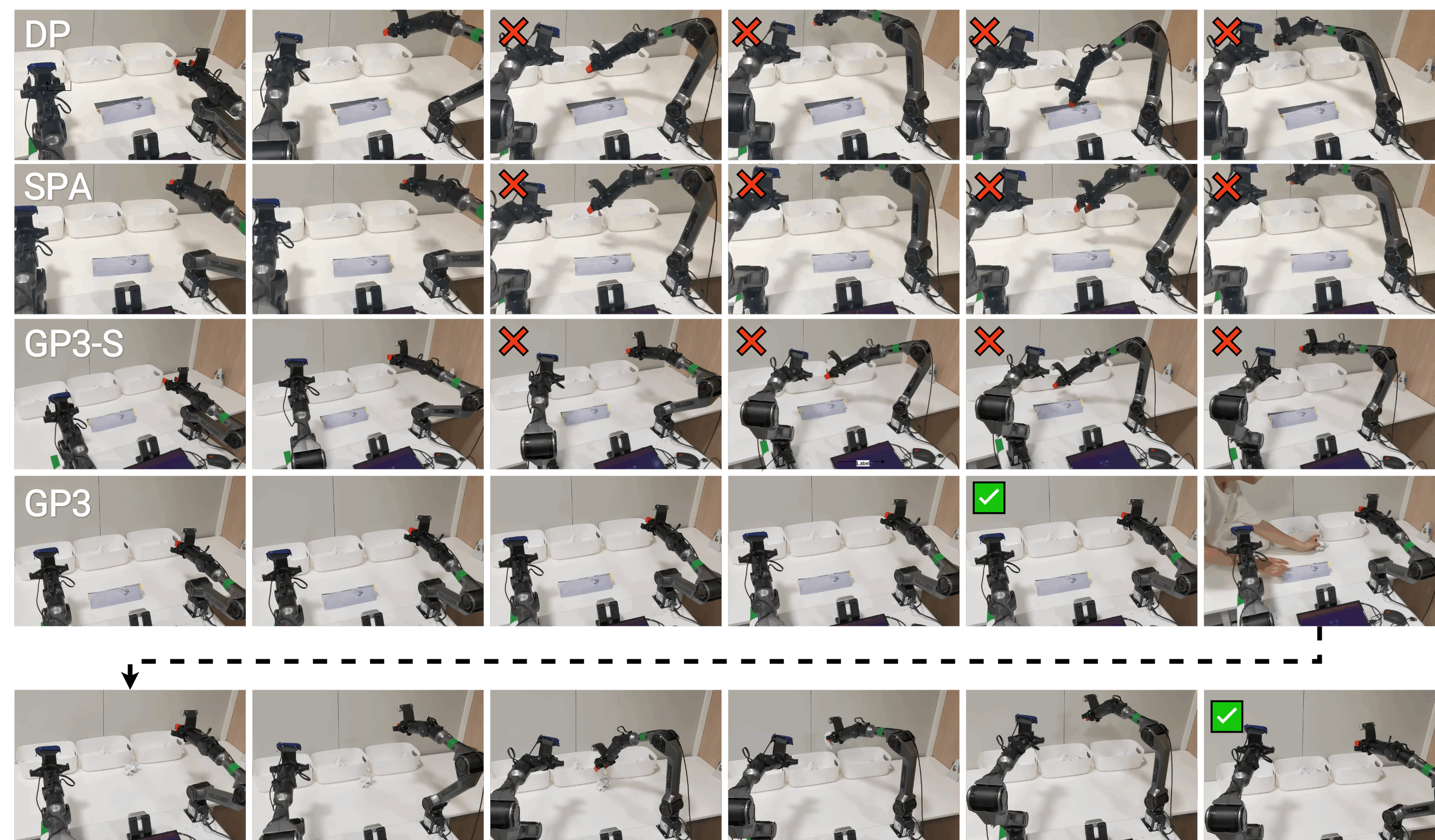}
    \caption{Examples of 3D-aware task executions. Only the multi-view input GP3 will not be deceived by printed flat paper.}
    \label{fig:gp3_3d_understanding}
\end{figure}

{\bfseries 3D-Aware Tasks. }We further design a challenging task that requires 3D spatial awareness to evaluate the 3D understanding capability of GP3. As shown in Figure~\ref{fig:gp3_3d_understanding}, we print a flat paper ball and place it on the desk, which appears visually identical to a normal paper ball from the front-facing camera. However, a model with robust 3D perception should be able to distinguish between the two based on multi-view information. As observed, the 1-view GP3-S, as well as the 3-view SPA and DP, are all deceived into grasping the flat object. In contrast, the 3-view GP3 correctly identifies the flat paper ball and does not attempt to grasp it. Moreover, when the flat paper ball is replaced with a normal one, GP3 successfully executes the grasping action. This experiment further demonstrates that GP3 possesses strong capabilities in perceiving and understanding 3D spatial relationships.

\label{apdx:sec:data_curation}

\end{document}